\documentclass{article}

\usepackage{ifthen}
\usepackage{tabulary}
\newboolean{arxiv}
\setboolean{arxiv}{true}
\usepackage{array} 

\usepackage{algorithm}
\usepackage{algorithmic}

\usepackage[authoryear,round]{natbib}

\usepackage{ifthen}

\ifthenelse{\boolean{arxiv}}{
\usepackage{arxiv_def}
}
{
\usepackage{times}
\usepackage{colt_def}
}

\usepackage[toc,page]{appendix}
\usepackage{xr}
\usepackage{mathtools}
\usepackage{yfonts}  
\usepackage{subcaption} 

\DeclareSymbolFont{Greekletters}{OT1}{iwona}{m}{n}
\DeclareSymbolFont{greekletters}{OML}{iwona}{m}{it}
\DeclareMathSymbol{\salpha}{\mathord}{greekletters}{"0B}
\DeclareMathSymbol{\sbeta}{\mathord}{greekletters}{"0C}
\DeclareMathSymbol{\sgamma}{\mathord}{greekletters}{"0D}
\DeclareMathSymbol{\sOmega}{\mathord}{Greekletters}{"0A}
\DeclareMathSymbol{\smu}{\mathord}{greekletters}{"16}
\DeclareMathSymbol{\svarepsilon}{\mathord}{greekletters}{"22}
\DeclareMathSymbol{\svarrho}{\mathord}{greekletters}{"25}
\DeclareMathSymbol{\svarphi}{\mathord}{greekletters}{"27}

\allowdisplaybreaks
\usepackage[top=1in,bottom=1in,left=1in,right=1in]{geometry}

\newcommand{\aj}[1]{\textcolor{blue}{[AJ: #1]}}

\begin{document}

\title{Train on Validation (ToV):\\
Fast data selection with applications to fine-tuning
 }


\author{Ayush Jain${}^*$\and  Andrea Montanari\thanks{Granica Computing Inc. --- \href{www.granica.ai}{granica.ai}}\ 
 \thanks{Department of Statistics and Department of Mathematics, Stanford University}\and 
Eren Sasoglu${}^*$}

\maketitle

\begin{abstract}
State-of-the-art machine learning  often follows a two-stage process: $(i)$~pre-training on large, general-purpose datasets; $(ii)$~fine-tuning on task-specific data.  In fine-tuning, selecting training examples that closely reflect the target distribution is crucial. However, it is often the case that only a few samples are available from the target distribution. Existing data selection methods treat these target samples as a validation set and estimate the effect of adding or removing a single sample from the training pool by performing inference on the validation set.

We propose a simpler and faster alternative that inverts the usual role of train and validation: we perform inference on the training pool before and after fine-tuning on the validation set. We then select samples whose predictions change the most. Our key insight is that the training samples most affected by fine-tuning on a small validation set tend to be the most beneficial for reducing test loss on the target distribution. Experiments on instruction tuning and named entity recognition tasks show that, in most cases, our method achieves lower test log-loss than state-of-the-art approaches. We support our findings with theoretical analysis.

\end{abstract}

\section{Introduction}
\label{sec:Introduction}

While large language models (LLMs) are pretrained on internet-scale datasets, their downstream
performance can be heavily dependent on the instruction-tuning stage in which they are
fine-tuned on instruction/output pairs \citep{ouyang2022training,zhou2024lima,longpre2023flan,chung2024scaling}.
These datasets are significantly smaller
and are often gathered by using multiple heterogeneous
sources. Instruction tuning becomes even  more difficult when targeting a specialized use case \citep{wang2023far}. 
More generally, scarcity of domain-specific data is a ubiquitous challenge when fine-tuning foundation models. 

This paper presents an easy-to-implement and low-complexity method for selecting a training dataset of  prescribed size from heterogeneous sources to maximize the test time performance on the target distribution.
Our method is motivated by the theory of influence functions \citep{van2000asymptotic} yet
avoids the computational burden of computing influence functions. 
We validate this approach on two token-based learning tasks,
instruction tuning and named entity recognition (NER), and show that in most cases it outperforms state-of-the-art data selection baselines. To illustrate its broad applicability,
we show that it yields interesting results even for a simple logistic regression example (see Appendix~\ref{app:logreg}).

To formalize the problem, assume access to two datasets: 
a small dataset from the target distribution
$\P$ on $\cZ$ and a larger one from possibly heterogeneous data sources. We refer
to the dataset from the target as the `validation set'
$\bZv :=(\bzv_1, \dots, \bzv_{m_{\sval}})$ 
where  $\bzv_i$ are i.i.d. samples from the target distribution $\P$,
and to the larger heterogeneous dataset as `training pool' 
$\bX= (\bx_1,\dots,\bx_N)$, where $\bx_i\in\cZ$.
In general the distribution of the training pool differs from $\P$.
Our goal is to minimize the test error on the target distribution
with respect to the model parameters $\btheta\in\R^p$:
\begin{align}
R(\btheta) := \E[\ell(\btheta,\bz)] \, ,
\end{align}
where $\ell:\R^p\times \cZ\to \reals$ is a loss function.
A separate target-distribution test set $\bZt $  (separate from $\bZv$) is used to estimate $R(\btheta)$ after fine-tuning.

We aim to achieve this by training (or fine-tuning) the model on a subset $S\subseteq [N]$ of the training pool, e.g.
running stochastic gradient descent (SGD) with respect to the empirical risk:
\begin{align}
\hR_S(\btheta) := \frac{1}{|S|}\sum_{i\in S}\ell(\btheta,\bx_i) \, .
\end{align}
Let $\hbtheta_S$ be the outcome of running SGD (or any specific training algorithm) on $\hR_S(\btheta)$. We want to select the subset $S$ (given a constraint on its size $|S|$) so that $\hbtheta_S$ achieves a small test loss on the target distribution,
i.e. as to minimize $R(\hbtheta_S)$.
%



\subsection{Train on validation: motivation and algorithm}
\label{sec:IntroOverview}

To select the most helpful examples at model $\btheta$, we might score training examples by the decrease in validation loss induced by a single gradient step with respect to that example, then select those with the highest scores. Computing these scores directly requires $N{+}1$ full evaluations over the validation set.  We derive an efficient approximation to these scores.

Consider a single gradient step with respect to a training example $\bx$:
\begin{align}
    \btheta_{\bx} = \btheta-\eta \nabla\ell(\btheta, \bx)\, .\label{eq:SingleStep}
\end{align}

The corresponding change in loss for a validation example $\bz$, $\ell(\btheta, \bz)-\ell(\btheta_{\bx}, \bz)$, can be approximated by a first-order Taylor expansion:
\begin{align}
    \ell(\btheta, \bz)-\ell(\btheta_{\bx}, \bz) \approx -\langle\nabla \ell(\btheta, \bz),  \btheta_{ \bx}-\btheta\rangle = \eta\langle\nabla \ell(\btheta, \bz),  \nabla \ell(\btheta, \bx)\rangle,
\end{align}
where the last step follows from Eq.~\eqref{eq:SingleStep}.
 \cite{pruthi2020estimating} approximate the scores by computing gradients for each training and validation example and taking their dot products; \cite{xia2024less} extend this to token-based learning. Our method diverges from these approaches: it requires no per-example gradients.
 
Note the right-hand side is symmetric in $\bx$ and $\bz$. In other words, the decrease in loss on $\bz$ from a step on $\bx$ is mirrored by the decrease in loss on $\bx$ from a step on $\bz$. Our method exploits this train–validation symmetry.
The change in overall validation loss for a single gradient step with respect to $\bx$ is:
\begin{align}
  \frac{1}{m_{\sval}}\sum_{i=1}^{m_{\sval}} \Big(\ell(\btheta, \bz_i)-\ell(\btheta_{\bx}, \bz_i)\Big) &\approx  \frac{1}{m_{\sval}}\sum_{i=1}^{m_{\sval}} \eta\langle\nabla \ell(\btheta, \bz_i),  \nabla \ell(\btheta, \bx)\rangle\label{eq:ChangeVal}.
\end{align}
On the other hand, performing a batch gradient step at $\btheta$ 
with respect to the validation set gives 
\[
\btheta_{\bZv} = \btheta- \eta\frac{1}{m_{\sval}}\sum_{i=1}^{m_{\sval}}\nabla\ell(\btheta, \bz_i)\, .\]
Combining the above equation with Eq.~\eqref{eq:ChangeVal}, we get
\begin{align}
   \frac{1}{m_{\sval}}\sum_{i=1}^{m_{\sval}} \Big(\ell(\btheta, \bz_i)-\ell(\btheta_{\bx}, \bz_i)\Big)
   &\approx  \langle\btheta- \btheta_{\bZv} , \nabla\ell(\btheta, \bx)\rangle \approx \ell(\btheta, \bx)-\ell(\btheta_{\bZv}, \bx)\, .
   \label{eq:BasicIdentity_0}
\end{align}
In other words, the change in average validation loss from training on $\bx$ can be approximated by the change in loss on $\bx$ after training on the validation set $\bZv$.

\begin{algorithm}[t]
\caption{ToV  Scoring Algorithm: \emph{Method A.} }
\label{alg:tov}
\begin{algorithmic}[1]
\STATE \textbf{Input:} Pretrained model $\btheta_0$, validation set $\bZv$,
training pool $\bX = (\bx_i : i \in [N])$, epochs $L$,
\STATE \phantom{Input:}\;\; selected data count $n$, learning-rate schedule $\{\eta_k\}_{k=1}^L$,  base model count $m$, $\eps\in [0, 1) $   
\STATE \textbf{Output:} Set of examples $S\subset [N]$ of size $n$ 
\STATE Sample base subset $U \subseteq [N]$ of size $m$ randomly; define $\bX_U = (\bx_i : i \in U)$
\STATE Initialize model: $\hbts_0 \gets \btheta_0$; set scores $\phi_i \gets 0$ for all $i \in [N] \setminus U$
\FOR{$k = 1$ to $L$}
    \STATE Train $\hbts_{k-1}$ on $\bX_U$ for one epoch with learning rate $\eta_k$ to obtain $\hbts_k$ 
    \STATE Train $\hbts_{k}$ for one epoch on $\bZv$ with a learning rate $\eps\eta_k$ to obtain $\hbtv_k$
    \FOR{each $i \in [N] \setminus U$}
        \STATE $\phi_i^{(k)} \gets F(\ell(\hbtv_k;\bx_i)-\ell(\hbts_k;\bx_i))$ (see Section~\ref{sec:Implementation} for the definition of $F$)
        \STATE $\phi_i \gets \phi_i + \phi_i^{(k)}/L$
    \ENDFOR
\ENDFOR
\STATE Return set $S\subseteq [N] \setminus U$ of size $n$ on the basis of scores $\phi_i$ (see text) 
\end{algorithmic}
\end{algorithm}

Our main objective is to evaluate the left-hand side of Eq.~\eqref{eq:BasicIdentity_0}
for all $\bx$ in the training set. The right-hand side provides a far more efficient route: $(i)$~Compute the loss $\ell(\btheta,\bx)$ 
for all training examples; $(ii)$~fine-tune $\btheta$ on the validation set to obtain $\btheta_{\bZv}$; $(iii)$~re-evaluate the new loss  $\ell(\btheta_{\bZv},\bx)$  at each training sample $\bx$, and approximate the effect of training on $\bx$ by computing the difference with the loss at point~$(i)$. 

This requires one epoch of training on the validation set and two evaluations over the training pool, as opposed to $N$ evaluations of the validation loss as suggested by a direct
evaluation of the left-hand side of Eq.~\eqref{eq:BasicIdentity_0}, and it does not require access to per-example gradients.

In the next sections we use this idea to obtain a selection algorithm
that alternates training on a
subset of the training set and on the validation set.
A specific implementation, which we refer to as `Method A', is given in Algorithm \ref{alg:tov};
a slightly different implementation (`Method B')  will be given in Algorithm \ref{alg:tovB}.
In Method A, we start with a small random subset $U\subset [N]$ of the training pool.
We train on $U$ for $L$ epochs, resulting in models 
$\hbts_{1}$, \dots $\hbts_{L}$. 
For each epoch $k\in [L]$ we fine-tune $\hbts_{k}$ for one epoch on the validation set,
resulting in models $\hbtv_{k}$. 
For each epoch, every remaining training example $\bx_i$ with $i\in [N]\setminus U$ is scored by the change in its loss between $\hbts_{k}$ and $\hbtv_{k}$, and scores are averaged across epochs.

After computing scores $\phi_i$ as in Algorithm~\ref{alg:tov}, we select $S$ using one of two strategies:
$(i)$ choose the $n$ examples with the largest $\phi_i$;
$(ii)$ choose half from the highest-scoring examples and the other half uniformly at random from $U$ to increase diversity.

Intuitively, large $\phi_i$
means that a small amount of training on the target distribution 
produces a large change in the model output at $\bx_i$.
Our working assumption, motivated by the heuristics above and formalized in Section~\ref{sec:Justification}, is that the converse also holds: training on $\bx_i$ will produce a large change in model output on the target distribution.
Hence the scores $\phi_i$ can be used to select `important' samples for the target.

An adaptation for token-based learning is described in Section \ref{sec:TokenBased}, along with empirical results.
Section \ref{sec:Justification} provides a mathematical justification that formalizes the argument above.

\subsection{Related work}
Our work relates to data selection and data attribution. 
The impact of a single example on the validation error can be approximated by a 
first-order
Taylor expansion. This idea results in  data selection methods based on influence functions
\citep{wang2018optimal,wang2020less,ai2021optimal,kolossovtowards}.
Classical influence functions estimate the effect of a single example on the empirical risk minimizer.
Most closely related to our work are  \cite{pruthi2020estimating,bae2024training,xia2024less},
which instead estimate the influence of an example on the training dynamics.
In particular, \cite{bae2024training} shows how to approximately
propagate gradient changes at $k$-th epoch through all  subsequent epochs.
In contrast,  \cite{pruthi2020estimating,xia2024less} make a crude approximation for this propagation.
Limitations of influence-based methods are discussed in \cite{schioppa2023theoretical}.

The recent work of \cite{xia2024less} proposes LESS, a data selection method for instruction tuning that adapts influence ideas to Adam and long sequences.  
In particular, these authors emphasize the challenge of computing and storing 
gradients to compute influences.  
They address this problem via random projections and low-rank approximation.
\cite{engstrom2024dsdm} apply the datamodel framework
\citep{ilyas2022datamodels,park2023trak} to select pretraining data. Separately, a \emph{replay} algorithm that stores only a logarithmic number of checkpoints is proposed in \cite{engstrom2025optimizing}.
Methods that align training data distributions to a small target set include TSDS \citep{liu2024tsds} and DSIR \citep{xie2023data}; domain/task-adaptive pretraining also improves transfer \citep{gururangan2020don}. Broader LLM data-efficiency work proposes LLM-guided quality scoring (Ask-LLM) and density sampling \citep{sachdeva2024train}, and clustering-based sensitivity sampling with provable guarantees \citep{axiotis2024data}.  
Finally, Data Filtering Networks (DFN) also leverage a held-out, high-quality set, but with a different goal and setup
\citep{fang2023data}.

Our contribution differs by $(i)$~inverting train/validation roles to approximate per-example influence using only forward losses and doesn't require per example gradients, or Hessian-vector products—and $(ii)$~showing that this simple, symmetry-based score is computationally inexpensive and outperforms recent data selection approaches for instruction tuning and NER.
%
%
\section{Data selection for token-based learning}
\label{sec:TokenBased}

In this section we describe our implementation of the general idea
described in the introduction for token-based learning and present empirical results demonstrating its effectiveness.
Since prediction takes place at the token level, while data selection takes place 
at the example level (e.g., instruction/output pair), we compute token scores and
aggregate them as described in Section 
\ref{sec:Implementation}. Section~\ref{sec:TaskOver} gives a brief overview of instruction-tuning and NER tasks. Experimental settings are introduced in
Section \ref{sec:Settings}. Empirical results are presented
in Sections \ref{sec:Experiments-IT} and \ref{sec:Experiments-NER}.

\subsection{Score computation for token-based learning}
\label{sec:Implementation}

Each example $\bz$ consists of an input  $\bz^{\sin}$ and an output $\bz^{\sout}$, both of which are strings and may differ in length. Let $\cZ^{\sout}$ denote the output vocabulary, and let $T(\bz)$ denote the length of the output string $\bz^{\sout}$, which we write as $\bz^{\sout} = \big(z^{\sout}(1), z^{\sout}(2), \dots, z^{\sout}(T(\bz))\big)$.

Given a model parameterized by $\btheta$, its prediction on example $\bz$ is a sequence of $T(\bz)$ conditional distributions, $\{p_t(\cdot \mid \bz, \btheta)\}_{t=1}^{T(\bz)}$, where each $p_t(\cdot \mid \bz, \btheta)$  denotes the model’s predictive distribution over the output token at position $t$. Note that $p_t(\cdot \mid \bz, \btheta)$ depends on $\bz$ solely through
$\bz^{\sin}$ and $z^{\sout}(1), \dots, z^{\sout}(t-1)$.
We train models using the log-loss
\begin{align} 
\ell(\btheta; \bz) = -\frac{1}{T(\bz)} \sum_{t=1}^{T(\bz)} \log p_t\big(z^{\sout}(t) \mid \bz; \btheta\big). 
\end{align}
To compare two models, $\btheta$ and $\btheta'$, on example $\bz$, we define a per-token difference of log-loss 
\begin{equation}
\Delta_t(\bz; \btheta, \btheta') = \log \frac{p_t\big(z^{\sout}(t) \mid \bz; \btheta'\big)}{p_t\big(z^{\sout}(t) \mid \bz; \btheta\big)}. \end{equation} 

Since our setting involves selecting entire examples rather than individual tokens, we aggregate the per-token differences into a single score per example. Specifically, we apply a transformation function $F: \reals \to \reals$ to each $\Delta_t$ before averaging across positions. The final score for example $\bz$ is: 
\begin{align} 
\phi(\bz; \btheta, \btheta') = \frac{1}{T(\bz)} \sum_{t=1}^{T(\bz)} F\big(\Delta_t(\bz; \btheta, \btheta')\big). 
\end{align}

We consider three instantiations of the function $F$, leading to three different scoring methods: 

\noindent {\sc Maximum-Improvement:} $F(y) = y$ — emphasizes raw improvement.

\noindent{\sc Maximum-Absolute Change:} $F(y) = |y|$ — captures the magnitude of change. 

\noindent {\sc Maximum-Positive Improvement:} $F(y) = \max\{y, 0\}$ — ignores degradations.

The  algorithm is therefore the same as in Algorithm \ref{alg:tov},
with the adaptation $\phi_i^{(k)} =\phi(\bx_i;\hbts_k,\hbtv_k)$.

Given a budget of $n$ examples, we choose $S\subseteq [N]\setminus U$, $|S|=n$
using one of these rules:

 \noindent {\sc Score-only:} Choose the $n$ examples $i\in [N]\setminus U$
that have the largest score $\phi_i$.

 \noindent  {\sc Score+random:} Choose the $n/2$ examples $i\in [N]\setminus U$
that have the largest score $\phi_i$, and add $n/2$ more examples chosen uniformly at random 
(without replacement) from 
$U$.

Our scoring schemes tend to favor shorter examples due to their higher variance, which arises from having fewer tokens. To mitigate this bias, we partition the set $[N] \setminus U$ into 10 bins based on sequence length, ensuring each bin contains an equal number of examples. We then select an equal number of top-scoring examples from each bin.


After selecting $S$ of size $|S|=n$,
we train (or fine tune) a  model on $S$ to evaluate the selection scheme. We refer to this stage as \emph{final training}.

We compare our schemes with three baselines:

\noindent{\sc Random:} The set $S$ is selected uniformly at random subject to its size.

\noindent {\sc Maximum uncertainty:} Instead of the scores we defined, we use the
following hardness score:
\begin{align}
\psi_i := \frac{1}{T_i}\sum_{t=1}^{T_i}\log\big(\sp_{t}(z_i(t) |\bz_i;\hbts_L) (1-
\sp_{t}(z_i(t) |\bz_i;\hbts_L)\big) \, ,
\end{align}
 This score extends the method of \cite{ting2018optimal,wang2018optimal,ai2021optimal,kolossovtowards}
 to token-based learning .

\noindent {\sc LESS:} We used the publicly available implementation from \cite{xia2024less}; see Appendix~\ref{app:tbl}.\looseness-1


\subsection{Prediction tasks}\label{sec:TaskOver}
We evaluate our data selection framework in two distinct token-based tasks: instruction tuning (IT) and named entity recognition (NER).
The framework we introduced above captures both tasks:

\noindent{\bf Instruction Tuning (IT)}
involves training a language model to follow natural language instructions. Each training example consists of:\\
\textbf{Input} $\bz^{\sin}$: a user instruction or prompt; \;\;\textbf{Output} $\bz^{\sout}$: the desired model response.

The output is typically multi-token and highly variable in content and length, depending on the instruction. The model learns to generate $\bz^{\sout}$ conditioned on $\bz^{\sin}$. This naturally fits our framework, which models predictions as token-level distributions $p_t(\cdot \mid \bz, \btheta)$.

\noindent{\bf Named Entity Recognition (NER)} is a sequence labeling task where the model assigns a probability distribution over entity tags (e.g., PERSON, ORGANIZATION, \dots) to each token. In this case:\\
\textbf{Input} $\bz^{\sin}$: a tokenized sentence;\;\;
   \textbf{Output} $\bz^{\sout}$: a sequence of entity labels, aligned with the input.
   
In NER, predictions are computed as token-wise classification distributions and therefore output is of the same length as input sequence.\footnote{In NER, typically token level probabilities are combined to assign labels to a whole word.}
In this case, as a base model we take a pretrained language model and replace its prediction head with a classification head.

\subsection{Experimental setting}
\label{sec:Settings}

In all of our experiments the training set consisted of $N=36\times 1024$ samples. For the base model training, we used $|U|=4\times 1024$ samples. The validation set size is $m_{\sval}=1024$ and the test set size is $m_{\stst} = 10,000$. We vary the selected 
set size $n\in\{1,2,4,8\}\times 1024$.

\noindent{\bf Number of epochs.} Both for surrogate model training and final model training
we determine the number of epochs by 
$L = (16\times 1024)/n_{\str}$.    
We use a batch size of $16$ whence the above ensures that the number batches used in training remains constant, and equal to $1024$. In other words, all experiments in this section are 
at \emph{constant compute}.
Since base model training uses $|U|=m=4\times 1024$ samples, the number of epochs 
is $L=4$.

\noindent{\bf Learning rate.} The learning rate for both surrogate and final model training is selected using hyper-parameter optimization for each selected set size $n$.
The learning-rate optimization was carried out for random data selection hence
placing our approach at a disadvantage.

We use linear learning rate scheduler and  LoRA training \cite{hu2022lora}
with LoRA parameters  $\alpha =32$ and  ${\sf dropout} =0.2$.
For NER experiments, we used ${\sf PEFT rank}=1$ and for instruction tuning experiments, we used
${\sf PEFT rank}=256$.
The learning rate for the validation examples is $\eps=1/10$ of the one for the base examples.
We present here results with {\sc Score$+$random} and refer to the appendix for
{\sc Score-only}.

\subsection{Experiments for instruction tuning}
\label{sec:Experiments-IT}

For these experiments we used 3 different datasets, which we will refer to as $\cuS := \{$Slim Orca, Alpaca GPT-4, Alpaca GPT-3.5$\}$. As the foundation model, we use \textbf{Meta-Llama-3-8B}. Additional details of the model and datasets used are provided in the Appendix.

We designed five experimental setups. In each experiment, one dataset from $\cuS$ is selected as the \emph{target distribution}. We randomly sample validation and test sets, $\bZv$ and $\bZt$, without replacement from the target dataset. These samples are excluded from further use.
The \emph{training pool} is then formed by randomly sampling an equal number of examples from one or more datasets in $\cuS$ (excluding the validation and test samples), such that the total number of selected training samples is fixed at $N$. We denote by $\cuS_*\subseteq \cuS$
the datasets used to generate the training pool. The choices of the target dataset and of $\cuS_*$ for each of the five experiments are summarized in Table~\ref{table:IT_exp}.
All reported results are averaged over 10 independent runs. In each run, we freshly sample the training, validation, and test sets. These experiments are designed to evaluate performance across a range of data configurations. In particular: in Experiments 1 and 4, the training set includes samples 
from both target distribution and other distributions;
in Experiments 2 and 5, the training set includes samples 
only from non-target distributions; in Experiment 3, it includes only samples from the
target distribution.\looseness-1

\begin{table}[h!]
\centering
\begin{minipage}{0.47\textwidth}
\centering
\vspace{1em}
\caption{Summary of instruction tuning experiments. Abbreviations: SO = Slim Orca, A4 = Alpaca GPT-4, A3.5 = Alpaca GPT-3.5.}
\vspace{1em}
\begin{tabular}{|c|c|p{0.55\linewidth}|}
\hline
\textbf{Exp} & \textbf{Target} & \textbf{Training pool} \\ \hline
1 & SO  & SO, A4, and A3.5 \\ \hline
2 & SO  & A4 and A3.5 \\ \hline
3 & SO  & SO \\ \hline
4 & A4  & SO, A4, and A3.5 \\ \hline
5 & A4  & SO and A3.5 \\ \hline
\end{tabular}
\vspace{1em}
\label{table:IT_exp}
\end{minipage}\hfill
\begin{minipage}{0.47\textwidth}
\centering
\vspace{1em}
\caption{Summary of named entity recognition experiments. Abbreviations: MN = Multinerd, Ai4p = Ai4p, C4 = C4, SB = Syn-big.}
\vspace{1em}
\begin{tabular}{|c|c|p{0.55\linewidth}|}
\hline
\textbf{Exp} & \textbf{Target} & \textbf{Training pool} \\ \hline
1 & MN   & MN, Ai4p, C4, and SB \\ \hline
2 & MN   & Ai4p, C4, and SB \\ \hline
3 & MN   & MN \\ \hline
4 & Ai4p  & MN, Ai4p, C4, and SB \\ \hline
5 & Ai4p  & MN, C4, and SB \\ \hline
6 & Ai4p  & Ai4p \\ \hline
\end{tabular}
\vspace{1em}
\label{table:NER_exp}
\end{minipage}
\end{table}

\begin{figure}[h!]
\begin{center}
    \includegraphics[width=\textwidth]
{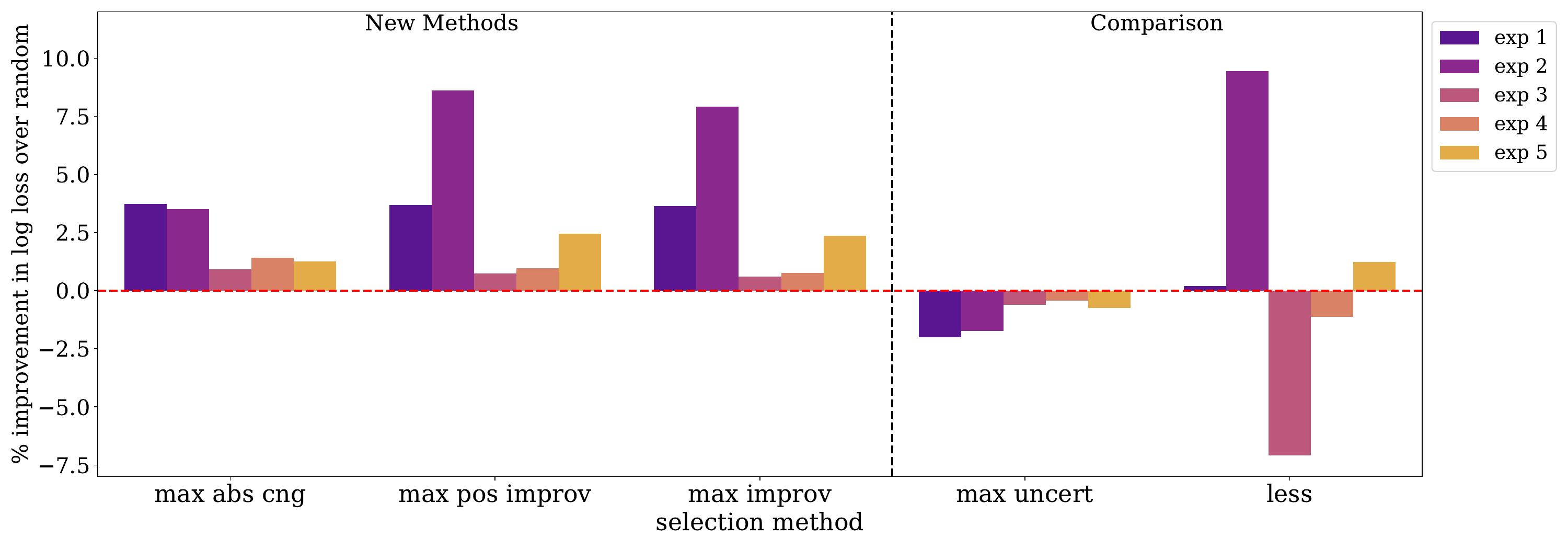}
\end{center}
\vspace{-0.75em}
    \caption{
Test log-loss improvement (\%) over random selection for instruction tuning with $n = 8 \times 1024$ samples.
Each group of bars represents a data-selection strategy (maximum-uncertainty and LESS as baselines); colors show target/training pool configuration (Table~\ref{table:IT_exp}).
Results use Method~A (Algorithm~\ref{alg:tov}) with the \textsc{Score+Random} strategy.
}\label{fig:IT_barplot}
\end{figure}

\begin{figure}[ht]
\begin{center}
\centering
\makebox[0.85\textwidth][2l]{%
    \includegraphics[trim=0 0 0 0,clip,width=1.15\textwidth]{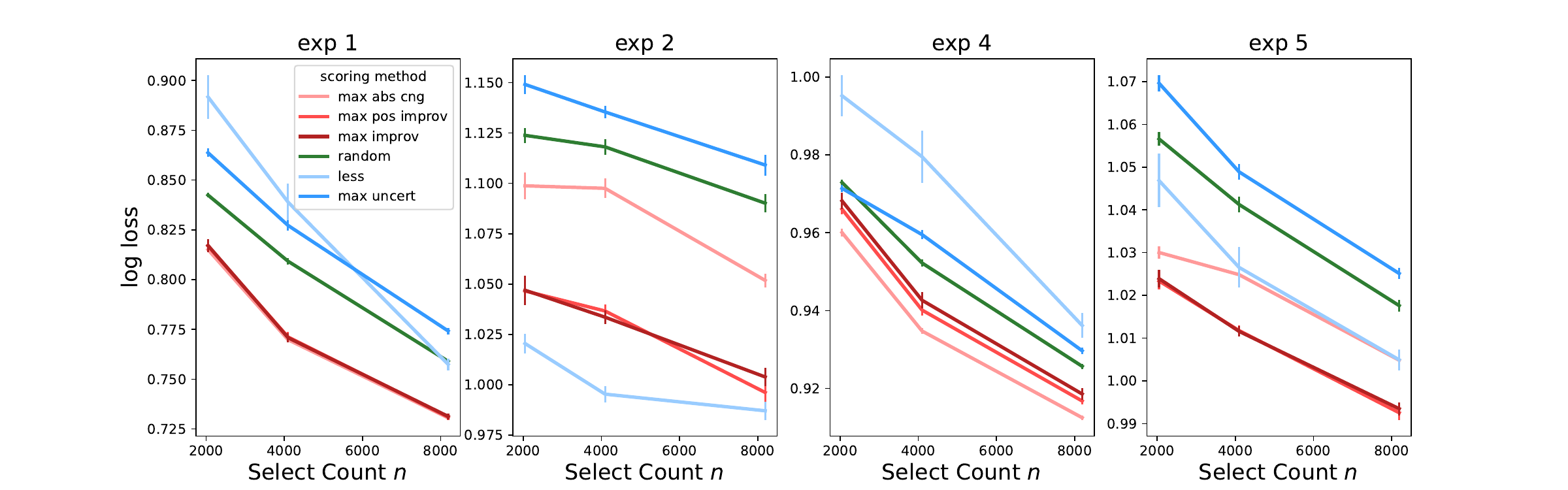}
}
\end{center}\vspace{-0.75em}

    \caption{
    Test log-loss vs. number of selected samples $n$ for instruction tuning.
(Due to space limits, Exp.~3 plot is in the Appendix.)
Lines show mean log-loss over 10 runs; error bars are $\pm$1 standard error.
Results use Method~A  with the \textsc{Score+Random} strategy.
}\label{fig:IT_evol}
\end{figure}

Figure \ref{fig:IT_barplot} summarizes our results for instruction tuning for a fixed select size $n=8\times 1024$. We plot the improvement in test log-loss over random data selection
for several data-selection strategies within the general framework
described
in Section \ref{sec:Implementation}, using method A in algorithm~\ref{alg:tov} for scoring the examples and {\sc score+random} for selecting.
 We observe that the proposed strategies yield significantly better instruction tuning than random data selection or selecting by max-uncertainty. We observe an improvement (albeit a small  one) even 
when both train and validation data are from Slim Orca (Exp 3), which is a case in which random selection should perform well. The proposed strategies also yield a significant improvement over LESS \citep{xia2024less},
with the exception of Experiment 2 in which LESS performs slightly  better.

Figure \ref{fig:IT_evol} displays the evolution of test log loss with selected sample size $n$. We observe that a good choice of the data selection method results in model improvements that can be equivalent to or larger than doubling $n$. Plots show standard error (with scaling factor 1) for 10 runs.

\begin{figure}[t]
\begin{center}
     \includegraphics[width=\textwidth]
    {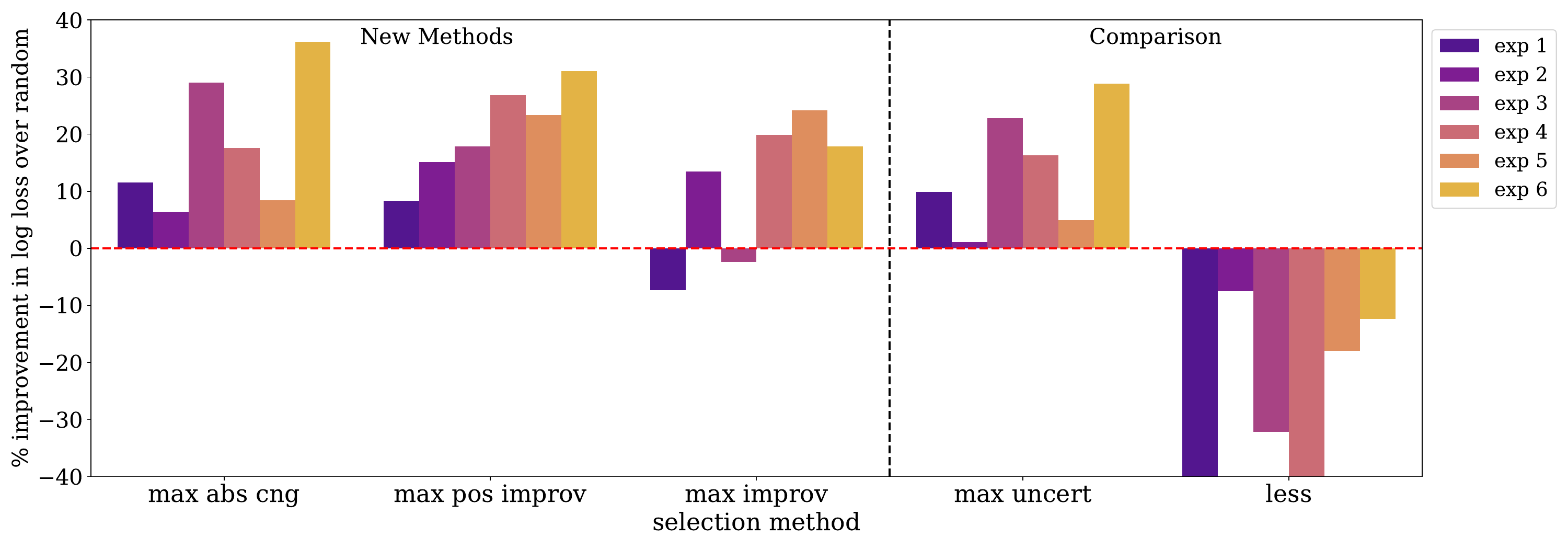}
\end{center}
\vspace{-0.75em}
    \caption{
Test log-loss improvement (\%) relative to random selection for NER at $n = 8 \times 1024$.
Each group of bars represents a data-selection strategy; colors show target/training pool configuration (Table~\ref{table:NER_exp}).
Results use Method~A (Algorithm~\ref{alg:tov}) with the \textsc{Score+Random} strategy}\label{fig:NER_barplot}
\end{figure}

\subsection{Experiments for named entity recognition}
\label{sec:Experiments-NER}

The task is to classify whether a
token is part of a person name or not.
For these experiments we used 4 different labeled datasets, 
which we will refer to as $\cuS:=\{$
Multinerd, Ai4p, C4, Syn-big$\}$.
 We use \textbf{xlm-roberta-base} as the foundation model.
Further details on the experiment, model and datasets used are presented in the Appendix.

We conducted six sets of experiments.
As for the case of instruction tuning, for each set of experiments, we select one of the datasets $\cuS$ as 
defining the target distribution, and one or more other datasets to define the training pool (denoted by $\cuS_*$). The choices of target datasets and $\cuS_*$ are summarized in 
Table \ref{table:NER_exp}. The construction of train, test and validation sets is same as in instruction tuning.\looseness-1

Figure \ref{fig:NER_barplot} summarizes our experiments with NER. 
We plot the improvement in test log-loss over random data selection
for several scores definitions.  Throughout these experiments, we use {\sc score+random}.
We observe that the strategies of Section \ref{sec:Implementation} yield 
systematic improvements over
random data-selection. Unlike in the case of instruction tuning, 
maximum uncertainty also yields an improvement in most settings.
However, the ToV approach achieves a larger improvement. 
Finally, in this case LESS \citep{xia2024less} appears not to improve over random data selection.
%
%
%
\section{A formal justification}
\label{sec:Justification}

In this section we present a mathematical
analysis of our approach in the case of
batch gradient descent (GD). 
We focus on the implementation \emph{Method B}, described in Algorithm \ref{alg:tovB}.

\begin{algorithm}[t]
\caption{ToV Scoring Algorithm: \emph{Method B} }
\label{alg:tovB}
\begin{algorithmic}[1]
\STATE \textbf{Input:} Pretrained model $\btheta_0$, validation set $\bZv$,
training pool $\bX = (\bx_i : i \in [N])$,
\STATE \phantom{Input:}\;\; selected data count $n\le N$, base model count $m$
\STATE \textbf{Output:} Set of examples $S\subset [N]$ of size $n$ 
\STATE Sample base subset $U \subseteq [N]$ of size $m$ randomly; define $\bX_U = (\bx_i : i \in U)$
\STATE Initialize models: $\hbtss_0 \gets \btheta_0$, $\hbts_0 \gets \btheta_0$; set scores $\Upsilon_i \gets 0$ for all $i \in [N] \setminus U$
\FOR{$k = 1$ to $L$}
\STATE  Train for one epoch on $\bX_U$ with learn. rate $\eta_k$ and init.
 $\hbtss_{k-1}$. Denote the output by $\hbtss_{0,k}$
 \STATE Train for one epoch on $\bZv$ with learn. rate $\eps\cdot\eta_k$ and init.
 $\hbtss_{0,k}$. Denote the output by $\hbtss_{k}$
 \STATE  Train for one epoch on $\bX_U$ with learn. rate $\eta_k$ and init.
 $\hbts_{k-1}$. Denote the output by $\hbts_{k}$
    \FOR{each $i \in [N] \setminus U$}
        \STATE $\Upsilon_i^{(k)} \gets \ell(\hbts_k; \bx_i)-\ell(\hbtss_k; \bx_i)$
        \STATE $\Upsilon_i \gets \Upsilon_i + \Upsilon_i^{(k)}/L$
    \ENDFOR
\ENDFOR
\STATE  Select $S\subseteq [N] \setminus U$ with size $|S|=n$ using scores $\Upsilon_i$
\end{algorithmic}
\end{algorithm}

Method B differs from Method A 
because at each training cycle $k$, training on the base set $\bX_U$
is initialized with the output of the previous train-on-validation phase. 
Empirically Method A performs somewhat better than B,
see Appendix \ref{sec:BvsA}. We use Method B for analysis just because the resulting mathematical expressions are simpler.

We find empirically that the ToV works well beyond token-based learning,
and hence our focus  will be to understand it in a generic learning problem.
 Appendix \ref{app:logreg} demonstrates this point by considering a 
 simple logistic regression  problem.

\subsection{Ideal scores, linearization, influence functions}

In order to estimate the model improvement produced by
sample $i\in [N]\setminus U$ we could train a model on two 
training sets $\bX_U$ and $\bX_{U\cup i}$, using empirical risk functions
$\hR_U(\btheta)$, $\hR_{U\cup i}(\btheta)$. We thus would run GD for $L$ steps,
with initialization  $\hbts_0=\hbtheta_{0}^{\ssur+i}=\btheta_0$:
\begin{align}
\hbtheta_{k+1}^{\ssur} = \hbtheta_{k}^{\ssur} -\eta m\nabla \hR_U(\hbtheta_{k}^{\ssur})\, ,
\;\;\;\;
\hbtheta_{k+1}^{\ssur+i} = \hbtheta_{k}^{\ssur+i} -\eta(m+1)\nabla \hR_{U\cup i}(\hbtheta_{k}^{\ssur+i})\, .\label{eq:TwoGD}
\end{align}
At iteration $k$, we have thus two models $\hbtheta_{k}^{\ssur}$
and $\hbtheta_{k}^{\ssur+i}$ that differ uniquely in whether sample $i$ is used or not. 
We define the \emph{ideal score} to be the difference in validation error between these two 
models, averaged over epochs 
\begin{align}
\score_i := \frac{1}{L}\sum_{s=1}^L[\hR_{\sval}(\hbtheta_{s}^{\ssur})-\hR_{\sval}(\hbtheta_{s}^{\ssur+i}) ]=
\frac{1}{m_{\sval} L}\sum_{s=1}^L\sum_{j=1}^{m_{\sval}} \big\{\ell(\hbtheta_{s}^{\ssur};\bzv_j)-\ell(\hbtheta_{s}^{\ssur+i};\bzv_j)\big\}\, .\label{eq:IdealScore}
\end{align}
Evaluating this score is computationally expensive, hence several groups 
\citep{pruthi2020estimating,bae2024training,xia2024less} proposed to use a first order Taylor expansion to approximate the difference
between the two models.
Expanding $\score_i$ with respect to 
the contribution of $\ell(\,\cdot\,;\bx_i)$ yields
\begin{align}
\score^{\slin}_i = \frac{\eta}{L}\sum_{0\le s<t\le L}
\<\nabla \hR_{\sval}(\hbtheta_{t}^{\ssur}),
\bM_{t,s+1}\nabla \ell(\hbtheta_{s}^{\ssur};\bx_i)\>\, .\label{eq:PsiLin}
\end{align}
where $\bM_{t,t}=\bI_d$ and $\bM_{t,r}$ captures the propagation
of perturbations along the GD trajectory:
\begin{align}
\bM_{t,r} &:=     \bH_{t-1}\cdot \bH_{t-2}\cdots \bH_{r}\, ,\;\;\;\;\;\;\;\;\;  \bH_k := \bI-\eta m
\nabla^2 \hR_U(\hbtheta_{k}^{\ssur})\, .\label{eq:MDef}
\end{align}
The next result shows that $\score^{\slin}_i$
approximates well  $\score_i$ in a quantitative way, under local convexity.
\begin{proposition}\label{propo:InfluenceFunctions}
Assume there exist $c_0, C_1, M>0$ such that $\nabla^2 \hR_U(\hbtheta_{k}^{\ssur})\succeq c_0\bI_d$,
$\|\nabla \ell(\hbtheta_{k}^{\ssur};\bx_i)\|\le C_1$
for all $k$ and, for all $\btheta_1,\btheta_2$, 
 $\|\nabla^2 \hR_U(\btheta_{1})-\nabla^2 \hR_U(\btheta_2)\|_{\op}\le M\|\btheta_1-\btheta_2\|_2$,  $\|\nabla \ell(\btheta_{1};\bx_i)-\nabla\ell (\btheta_2;\bx_i)\|_{\op}\le M\|\btheta_1-\btheta_2\|_2$.
 Further assume that $\|\nabla^2 \hR_{\sval}(\hbtheta_{k}^{\ssur})\|_{\op}\le C_1$
 and  $\|\nabla^2 \hR_{\sval}(\btheta_{1})-\nabla^2 \hR_{\sval}(\btheta_2)\|_{\op}\le M\|\btheta_1-\btheta_2\|_2$ 
 for all $\btheta_1,\btheta_2$ as well. Finally, assume there exists a constant $C_{\eta}$ such that $\eta_k=\eta\le C_{\eta}/m$ $\forall k$.
Then there exists $C= C(c_0,C_1,C_{\eta},M)$ such that
\begin{align}
\big|\score_i-\score_i^{\slin}\big|\le C/m^2\, .\label{eq:PropoIF}
\end{align}
\end{proposition}
The assumption $\eta\le C_{\eta}/m$ is justified by the fact that we expect the Hessian of
$\hR_U(\, \cdot\, )$ to be of order one, and hence the stepsize for this objective 
(which is given by $\eta m$ see Eq.~\eqref{eq:TwoGD}) should be of order one. As shown in the proof, 
the typical size of $\score_i^{\slin}$ is of order $1/m$, and hence Eq.~\eqref{eq:PropoIF}
establishes that the difference $|\score_i-\score_i^{\slin}|$ is negligible.



\subsection{Train-validation duality}

We consider Methods A and  B defined in Algorithms \ref{alg:tov}, \ref{alg:tovB}. 
We emphasize the dependence on $\eps$ by writing $\phi_i = \phi_i(\eps)$ and $\Upsilon_i=\Upsilon_i(\eps)$. It is easy to derive the small $\eps$ asymptotics
$\phi_i= \phi^{\slin}_i \eps+o(\eps)$, 
$\Upsilon_i= \Upsilon^{\slin}_i \eps+o(\eps)$,
where, for $\bg_{s,i}:=
\nabla \ell(\hbtheta_{s}^{\ssur};\bx_i)$,
\begin{align}
\phi^{\slin}_i := \frac{\eta\mval}{L}\sum_{s=1}^L
\<\nabla \hR_{\sval}(\hbtheta_{s}^{\ssur}),\bg_{s,i}\>\, ,\, \;\;\;\;\;
 \Upsilon^{\slin}_i := \frac{\eta\mval}{L}\sum_{0\le t<s\le L}
\<\nabla \hR_{\sval}(\hbtheta_{t+1}^{\ssur}),
\bM_{s,t+1}^{\sT}\bg_{s,i}\>\, .
\label{eq:PhiLin}
\end{align}
We show that these are good approximations 
of $\Upsilon_i(\eps), \phi_i(\eps)$ 
uniformly in dimension, sample size.
\begin{theorem}\label{propo:Score}
Consider Algorithms \ref{alg:tov}, \ref{alg:tovB} with fixed stepsize $\eta_k=\eta$
(and $F(x) = -x$  in Algorithm \ref{alg:tov}).
Under the assumptions of Proposition \ref{propo:InfluenceFunctions}, further assume $\|\nabla\hR_{\sval}(\hbts_k)\|\le C_1$ for all $k$, and  $\|\nabla^2_{\btheta}\ell(\btheta;\bx)\|_{\op}\le C_1$. Then there exist $c_*=c_*(c_0,M,C_1)$,
$C= C(c_0,M, C_1)$ such that, for $\eps m_{\sval}/m\le c_*$,
\begin{align}
\big|\Upsilon_i(\eps)-\Upsilon^{\slin}_i\eps\big|\le C\big(\eps m_{\sval}/m\big)^2\, ,\;\;\;\;
\big|\phi_i(\eps)-\phi^{\slin}_i\eps|\le C\big(\eps m_{\sval}/m\big)^2\, .
\end{align}
\end{theorem}
Note that $\Upsilon^{\slin}_i$ differ from $\score^{\slin}_i$.
because of: $(i)$~The different order of $s$ and $t$; $(ii)$~The fact that 
$\bM_{t,s+1}$ is replaced by its transpose in Eq.~\eqref{eq:PhiLin}.
$\Upsilon^{\slin}_i$ 
measures the influence
of training on validation data when making inference at $\bx_i$,
while  $\score^{\slin}_i$  measures the influence
of training on $\bx_i$ data when making inference on validation.
These two measures of `influence'  
differ by the  replacement of $\bM_{t,s+1}$ by  $\bM_{s,t}^{\sT}$.
However, in a number of cases we expect these two matrices to be 
not too different, and hence the two scores to yield similar results.
We can prove that $\Upsilon^{\slin}_i$ and $\score^{\slin}_i$ coincide 
(for large $L$) under local convexity conditions.
\begin{theorem}\label{thm:LargeL_cvx}
Assume $\btheta\mapsto \ell(\btheta;\bx)$ to be twice continuously differentiable 
and that $\|\nabla \hR_{\sval}(\hbts_k)\|\le C_1$, $\|\nabla \ell(\hbts_k;\bx_i)\|\le C_1$
for all $k$.
Further assume that gradient descent iterates $(\hbtheta^{\ssur}_{k}:k\ge 0)$
converge to $\hbtheta^{\ssur}_{\infty}=\lim_{k\to\infty}\hbtheta^{\ssur}_{k}$
which is a local minimum of $\hR_U(\btheta)$ with 
$\bQ_{\infty}:=\nabla^2\hR_U(\hbtheta^{\ssur}_{\infty})\succ \bzero$
 (strictly). 
Then 
\begin{align}\label{eq:InfluenceLocConvex}
\lim_{L\to\infty}\frac{1}{\mval}\Upsilon^{\slin}_i(L) = \lim_{L\to\infty} \score^{\slin}_i(L) = 
\frac{1}{m} \<\nabla \hR_{\sval}(\hbtheta_{\infty}^{\ssur}),
\bQ_{\infty}^{-1}\nabla \ell(\hbtheta_{\infty}^{\ssur};\bx_i)\>:= \score^{\slin}_{i,\infty}\, .
\end{align}
\end{theorem}
The last expression in Eq.~\eqref{eq:InfluenceLocConvex} (denoted by $\score^{\slin}_{i,\infty}$)
is the classical formula for influence functions of M-estimators \citep{van2000asymptotic}.
Both our approach and the  dynamical influence function $\score^{\slin}_i(L)$ can be regarded 
as approximations of $\score^{\slin}_{i,\infty}$ in this case.

In  fine tuning, the model is likely to be overparametrized,
and it is unrealistic to assume convergence to a strict minimum (with 
$\nabla^2\hR_U(\hbtheta^{\ssur}_{\infty})\succ \bzero$).
On the other hand, the weights will not change significantly during this phase
and it is reasonable to approximate fine-tuning as fitting an overparametrized 
linear  model with respect to the empirical neural tangent features learnt in the
pre-training phase. 
\begin{theorem}\label{thm:LargeL-Linear}
Consider the loss function $\ell(\btheta;\bx) = (y(\bx)- \<\bpsi(\bx),\btheta\>)^2/2$ for some response
variables $y(\bx)$, and featurization map $\bpsi:\reals^d\to \reals^p$, $p>m$.
Let $\bPsi\in\reals^{|U|\times p}$ be the matrix with rows $(\bpsi(\bx_j):j\in U)$,
$\bPsi_{\sval}\in\reals^{m_{\sval}\times p}$ be the matrix with rows $(\bpsi(\bz^{\sval}_j):j\le m_{\sval})$, $\bP_{\Psi}$
the projector to the kernel of $\bPsi$,
$\by = (y(\bx_j):j\in U)$, 
$\hbtheta:= \bPsi^{\dagger}\by$, $\br^{\sval}:= (y(\bz^{\sval}_j) - \<\hbtheta,\bpsi(\bz^{\sval}_j)\>:j\le m_{\sval})$,
$r(i):= y(\bx_i) -\<\hbtheta, \bpsi(\bx_i)\>$.
If GD is initialized with $\btheta_0=\bzero$, and we use constant stepsize $\eta<\|\bPsi\|^2_{op}/2$,
then
\begin{align}
 \lim_{L\to\infty}\frac{1}{L\mval}\Upsilon^{\slin}_i(L) =   \lim_{L\to\infty}\frac{1}{L}\score^{\slin}_i(L) =
\frac{\eta}{2} \, r(i)
\<\br^{\sval},\bPsi_{\sval}^{\sT}\bP_{\Psi} \bpsi(\bx_i)\>\, .\label{eq:LimitOverparam}
\end{align}
\end{theorem}

\newpage
\section*{Acknowledgements}
We are grateful to Neeraja Abhyankar, Alankrita Bhatt, Joseph Gardi,  Mukur Gupta, Germain Kolossov, Marc Laugharn, Rahul Ponnala, Sahasrajit Sarmasarkar, Andreas Santucci and Pulkit Tandon, for several conversations about this work.


%
 %

\bibliographystyle{plainnat}   
\bibliography{all-bibliography}

\begin{thebibliography}{34}
\providecommand{\natexlab}[1]{#1}
\providecommand{\url}[1]{\texttt{#1}}
\expandafter\ifx\csname urlstyle\endcsname\relax
  \providecommand{\doi}[1]{doi: #1}\else
  \providecommand{\doi}{doi: \begingroup \urlstyle{rm}\Url}\fi

\bibitem[Ai et~al.(2021)Ai, Yu, Zhang, and Wang]{ai2021optimal}
Mingyao Ai, Jun Yu, Huiming Zhang, and HaiYing Wang.
\newblock Optimal subsampling algorithms for big data regressions.
\newblock \emph{Statistica Sinica}, 31\penalty0 (2):\penalty0 749--772, 2021.

\bibitem[AI@Meta(2024)]{llama3modelcard}
AI@Meta.
\newblock Llama 3 model card.
\newblock 2024.
\newblock URL \url{https://github.com/meta-llama/llama3/blob/main/MODEL_CARD.md}.

\bibitem[Axiotis et~al.(2024)Axiotis, Cohen-Addad, Henzinger, Jerome, Mirrokni, Saulpic, Woodruff, and Wunder]{axiotis2024data}
Kyriakos Axiotis, Vincent Cohen-Addad, Monika Henzinger, Sammy Jerome, Vahab Mirrokni, David Saulpic, David Woodruff, and Michael Wunder.
\newblock Data-efficient learning via clustering-based sensitivity sampling: Foundation models and beyond.
\newblock \emph{arXiv preprint arXiv:2402.17327}, 2024.

\bibitem[Bae et~al.(2024)Bae, Lin, Lorraine, and Grosse]{bae2024training}
Juhan Bae, Wu~Lin, Jonathan Lorraine, and Roger Grosse.
\newblock Training data attribution via approximate unrolled differentation.
\newblock \emph{arXiv:2405.12186}, 2024.

\bibitem[Bartlett et~al.(2021)Bartlett, Montanari, and Rakhlin]{bartlett2021deep}
Peter~L Bartlett, Andrea Montanari, and Alexander Rakhlin.
\newblock Deep learning: a statistical viewpoint.
\newblock \emph{Acta numerica}, 30:\penalty0 87--201, 2021.

\bibitem[Chung et~al.(2024)Chung, Hou, Longpre, Zoph, Tay, Fedus, Li, Wang, Dehghani, Brahma, et~al.]{chung2024scaling}
Hyung~Won Chung, Le~Hou, Shayne Longpre, Barret Zoph, Yi~Tay, William Fedus, Yunxuan Li, Xuezhi Wang, Mostafa Dehghani, Siddhartha Brahma, et~al.
\newblock Scaling instruction-finetuned language models.
\newblock \emph{Journal of Machine Learning Research}, 25\penalty0 (70):\penalty0 1--53, 2024.

\bibitem[Conneau et~al.(2019)Conneau, Khandelwal, Goyal, Chaudhary, Wenzek, Guzm{\'{a}}n, Grave, Ott, Zettlemoyer, and Stoyanov]{DBLP:journals/corr/abs-1911-02116}
Alexis Conneau, Kartikay Khandelwal, Naman Goyal, Vishrav Chaudhary, Guillaume Wenzek, Francisco Guzm{\'{a}}n, Edouard Grave, Myle Ott, Luke Zettlemoyer, and Veselin Stoyanov.
\newblock Unsupervised cross-lingual representation learning at scale.
\newblock \emph{CoRR}, abs/1911.02116, 2019.
\newblock URL \url{http://arxiv.org/abs/1911.02116}.

\bibitem[Engstrom et~al.(2024)Engstrom, Feldmann, and Madry]{engstrom2024dsdm}
Logan Engstrom, Axel Feldmann, and Aleksander Madry.
\newblock Dsdm: Model-aware dataset selection with datamodels.
\newblock \emph{arXiv preprint arXiv:2401.12926}, 2024.

\bibitem[Engstrom et~al.(2025)Engstrom, Ilyas, Chen, Feldmann, Moses, and Madry]{engstrom2025optimizing}
Logan Engstrom, Andrew Ilyas, Benjamin Chen, Axel Feldmann, William Moses, and Aleksander Madry.
\newblock Optimizing ml training with metagradient descent.
\newblock \emph{arXiv preprint arXiv:2503.13751}, 2025.

\bibitem[Fang et~al.(2023)Fang, Jose, Jain, Schmidt, Toshev, and Shankar]{fang2023data}
Alex Fang, Albin~Madappally Jose, Amit Jain, Ludwig Schmidt, Alexander Toshev, and Vaishaal Shankar.
\newblock Data filtering networks.
\newblock \emph{arXiv preprint arXiv:2309.17425}, 2023.

\bibitem[Gururangan et~al.(2020)Gururangan, Marasovi{\'c}, Swayamdipta, Lo, Beltagy, Downey, and Smith]{gururangan2020don}
Suchin Gururangan, Ana Marasovi{\'c}, Swabha Swayamdipta, Kyle Lo, Iz~Beltagy, Doug Downey, and Noah~A Smith.
\newblock Don't stop pretraining: Adapt language models to domains and tasks.
\newblock \emph{arXiv preprint arXiv:2004.10964}, 2020.

\bibitem[Hu et~al.(2022)Hu, Wallis, Allen-Zhu, Li, Wang, Wang, Chen, et~al.]{hu2022lora}
Edward~J Hu, Phillip Wallis, Zeyuan Allen-Zhu, Yuanzhi Li, Shean Wang, Lu~Wang, Weizhu Chen, et~al.
\newblock Lora: Low-rank adaptation of large language models.
\newblock In \emph{International Conference on Learning Representations}, 2022.

\bibitem[Ilyas et~al.(2022)Ilyas, Park, Engstrom, Leclerc, and Madry]{ilyas2022datamodels}
Andrew Ilyas, Sung~Min Park, Logan Engstrom, Guillaume Leclerc, and Aleksander Madry.
\newblock Datamodels: Predicting predictions from training data.
\newblock In \emph{Proceedings of the 39th International Conference on Machine Learning}, 2022.

\bibitem[Kolossov et~al.(2024)Kolossov, Montanari, and Tandon]{kolossovtowards}
Germain Kolossov, Andrea Montanari, and Pulkit Tandon.
\newblock Towards a statistical theory of data selection under weak supervision.
\newblock In \emph{The Twelfth International Conference on Learning Representations}, 2024.

\bibitem[Lian et~al.(2023)Lian, Wang, Goodson, Pentland, Cook, Vong, and "Teknium"]{SlimOrca}
Wing Lian, Guan Wang, Bleys Goodson, Eugene Pentland, Austin Cook, Chanvichet Vong, and "Teknium".
\newblock Slimorca: An open dataset of gpt-4 augmented flan reasoning traces, with verification, 2023.
\newblock URL \url{https://https://huggingface.co/Open-Orca/SlimOrca}.

\bibitem[Liu et~al.(2024)Liu, Karbasi, and Rekatsinas]{liu2024tsds}
Zifan Liu, Amin Karbasi, and Theodoros Rekatsinas.
\newblock Tsds: Data selection for task-specific model finetuning.
\newblock \emph{Advances in Neural Information Processing Systems}, 37:\penalty0 10117--10147, 2024.

\bibitem[Longpre et~al.(2023)Longpre, Hou, Vu, Webson, Chung, Tay, Zhou, Le, Zoph, Wei, et~al.]{longpre2023flan}
Shayne Longpre, Le~Hou, Tu~Vu, Albert Webson, Hyung~Won Chung, Yi~Tay, Denny Zhou, Quoc~V Le, Barret Zoph, Jason Wei, et~al.
\newblock The flan collection: Designing data and methods for effective instruction tuning.
\newblock In \emph{International Conference on Machine Learning}, pages 22631--22648. PMLR, 2023.

\bibitem[Mukherjee et~al.(2023)Mukherjee, Mitra, Jawahar, Agarwal, Palangi, and Awadallah]{mukherjee2023orca}
Subhabrata Mukherjee, Arindam Mitra, Ganesh Jawahar, Sahaj Agarwal, Hamid Palangi, and Ahmed Awadallah.
\newblock Orca: Progressive learning from complex explanation traces of gpt-4, 2023.

\bibitem[Ouyang et~al.(2022)Ouyang, Wu, Jiang, Almeida, Wainwright, Mishkin, Zhang, Agarwal, Slama, Ray, et~al.]{ouyang2022training}
Long Ouyang, Jeffrey Wu, Xu~Jiang, Diogo Almeida, Carroll Wainwright, Pamela Mishkin, Chong Zhang, Sandhini Agarwal, Katarina Slama, Alex Ray, et~al.
\newblock Training language models to follow instructions with human feedback.
\newblock \emph{Advances in neural information processing systems}, 35:\penalty0 27730--27744, 2022.

\bibitem[Park et~al.(2023)Park, Georgiev, Ilyas, Leclerc, and Madry]{park2023trak}
Sung~Min Park, Kristian Georgiev, Andrew Ilyas, Guillaume Leclerc, and Aleksander Madry.
\newblock Trak: Attributing model behavior at scale.
\newblock \emph{arXiv preprint arXiv:2303.14186}, 2023.

\bibitem[Peng et~al.(2023)Peng, Li, He, Galley, and Gao]{peng2023instruction}
Baolin Peng, Chunyuan Li, Pengcheng He, Michel Galley, and Jianfeng Gao.
\newblock Instruction tuning with gpt-4.
\newblock \emph{arXiv preprint arXiv:2304.03277}, 2023.

\bibitem[Pruthi et~al.(2020)Pruthi, Liu, Kale, and Sundararajan]{pruthi2020estimating}
Garima Pruthi, Frederick Liu, Satyen Kale, and Mukund Sundararajan.
\newblock Estimating training data influence by tracing gradient descent.
\newblock \emph{Advances in Neural Information Processing Systems}, 33:\penalty0 19920--19930, 2020.

\bibitem[Sachdeva et~al.(2024)Sachdeva, Coleman, Kang, Ni, Hong, Chi, Caverlee, McAuley, and Cheng]{sachdeva2024train}
Noveen Sachdeva, Benjamin Coleman, Wang-Cheng Kang, Jianmo Ni, Lichan Hong, Ed~H Chi, James Caverlee, Julian McAuley, and Derek~Zhiyuan Cheng.
\newblock How to train data-efficient llms.
\newblock \emph{arXiv preprint arXiv:2402.09668}, 2024.

\bibitem[Schioppa et~al.(2023)Schioppa, Filippova, Titov, and Zablotskaia]{schioppa2023theoretical}
Andrea Schioppa, Katja Filippova, Ivan Titov, and Polina Zablotskaia.
\newblock Theoretical and practical perspectives on what influence functions do.
\newblock \emph{Advances in Neural Information Processing Systems}, 36:\penalty0 27560--27581, 2023.

\bibitem[Taori et~al.(2023)Taori, Gulrajani, Zhang, Dubois, Li, Guestrin, Liang, and Hashimoto]{alpaca}
Rohan Taori, Ishaan Gulrajani, Tianyi Zhang, Yann Dubois, Xuechen Li, Carlos Guestrin, Percy Liang, and Tatsunori~B. Hashimoto.
\newblock Stanford alpaca: An instruction-following llama model.
\newblock \url{https://github.com/tatsu-lab/stanford_alpaca}, 2023.

\bibitem[Tedeschi and Navigli(2022)]{tedeschi-navigli-2022-multinerd}
Simone Tedeschi and Roberto Navigli.
\newblock {M}ulti{NERD}: A multilingual, multi-genre and fine-grained dataset for named entity recognition (and disambiguation).
\newblock In \emph{Findings of the Association for Computational Linguistics: NAACL 2022}, pages 801--812, Seattle, United States, July 2022. Association for Computational Linguistics.
\newblock \doi{10.18653/v1/2022.findings-naacl.60}.
\newblock URL \url{https://aclanthology.org/2022.findings-naacl.60}.

\bibitem[Ting and Brochu(2018)]{ting2018optimal}
Daniel Ting and Eric Brochu.
\newblock Optimal subsampling with influence functions.
\newblock \emph{Advances in neural information processing systems}, 31, 2018.

\bibitem[van~der Vaart(2000)]{van2000asymptotic}
Aaad~W van~der Vaart.
\newblock \emph{Asymptotic Statistics}.
\newblock Cambridge University Press, 2000.

\bibitem[Wang et~al.(2018)Wang, Zhu, and Ma]{wang2018optimal}
HaiYing Wang, Rong Zhu, and Ping Ma.
\newblock Optimal subsampling for large sample logistic regression.
\newblock \emph{Journal of the American Statistical Association}, 113\penalty0 (522):\penalty0 829--844, 2018.

\bibitem[Wang et~al.(2023)Wang, Ivison, Dasigi, Hessel, Khot, Chandu, Wadden, MacMillan, Smith, Beltagy, et~al.]{wang2023far}
Yizhong Wang, Hamish Ivison, Pradeep Dasigi, Jack Hessel, Tushar Khot, Khyathi Chandu, David Wadden, Kelsey MacMillan, Noah~A Smith, Iz~Beltagy, et~al.
\newblock How far can camels go? exploring the state of instruction tuning on open resources.
\newblock \emph{Advances in Neural Information Processing Systems}, 36:\penalty0 74764--74786, 2023.

\bibitem[Wang et~al.(2020)Wang, Zhu, Dong, He, and Huang]{wang2020less}
Zifeng Wang, Hong Zhu, Zhenhua Dong, Xiuqiang He, and Shao-Lun Huang.
\newblock Less is better: Unweighted data subsampling via influence function.
\newblock In \emph{Proceedings of the AAAI Conference on Artificial Intelligence}, volume~34, pages 6340--6347, 2020.

\bibitem[Xia et~al.(2024)Xia, Malladi, Gururangan, Arora, and Chen]{xia2024less}
Mengzhou Xia, Sadhika Malladi, Suchin Gururangan, Sanjeev Arora, and Danqi Chen.
\newblock {LESS: selecting influential data for targeted instruction tuning}.
\newblock In \emph{Proceedings of the 41st International Conference on Machine Learning}, pages 54104--54132, 2024.

\bibitem[Xie et~al.(2023)Xie, Santurkar, Ma, and Liang]{xie2023data}
Sang~Michael Xie, Shibani Santurkar, Tengyu Ma, and Percy~S Liang.
\newblock Data selection for language models via importance resampling.
\newblock \emph{Advances in Neural Information Processing Systems}, 36:\penalty0 34201--34227, 2023.

\bibitem[Zhou et~al.(2024)Zhou, Liu, Xu, Iyer, Sun, Mao, Ma, Efrat, Yu, Yu, et~al.]{zhou2024lima}
Chunting Zhou, Pengfei Liu, Puxin Xu, Srinivasan Iyer, Jiao Sun, Yuning Mao, Xuezhe Ma, Avia Efrat, Ping Yu, Lili Yu, et~al.
\newblock Lima: Less is more for alignment.
\newblock \emph{Advances in Neural Information Processing Systems}, 36, 2024.

\end{thebibliography}

\newpage

\appendix

\section{Additional Experimental Details and Results}\label{app:addexp}

\subsection{Experiments for Token-Based Learning}\label{app:tbl}
In these experiments, we used pretrained models as base models and constructed training, validation, and test sets from real-world datasets. Details of the datasets and models are provided in Section~\ref{sec:datasets}. 

For each training example count—both for surrogate model training (used for scoring) and final model training—we selected the learning rate from the following grid:
\[
\texttt{[3e-6, 1e-5, 3e-5, 1e-4, 3e-4, 1e-3, 3e-3, 1e-2]}.
\]
The optimal learning rate was determined by training models on randomly sampled subsets from the training pool and evaluating their test log-loss. For each learning rate, the loss was averaged over 10 runs, with a new random subset used in each run. The best-performing learning rate was selected separately for each experimental configuration listed in Table~\ref{table:IT_exp} and Table~\ref{table:NER_exp}.

\noindent{\bf Implementation details for Less~\citep{xia2024less}}
We used the public implementation from the authors’ GitHub repository. The projection dimension was set to 8192. Learning rate and other hyperparameters were tuned identically for all approaches. For both our method and LESS, the surrogate model used the same number of samples and was trained for four epochs, matching the settings in the LESS paper. Following the original LESS procedure, we selected the top-scoring examples.

\subsection{Expanded Results for Instruction Tuning}

\begin{figure}[!htb]
\begin{center}
    \includegraphics[trim=50 20 0 20,width=1.1\textwidth]
{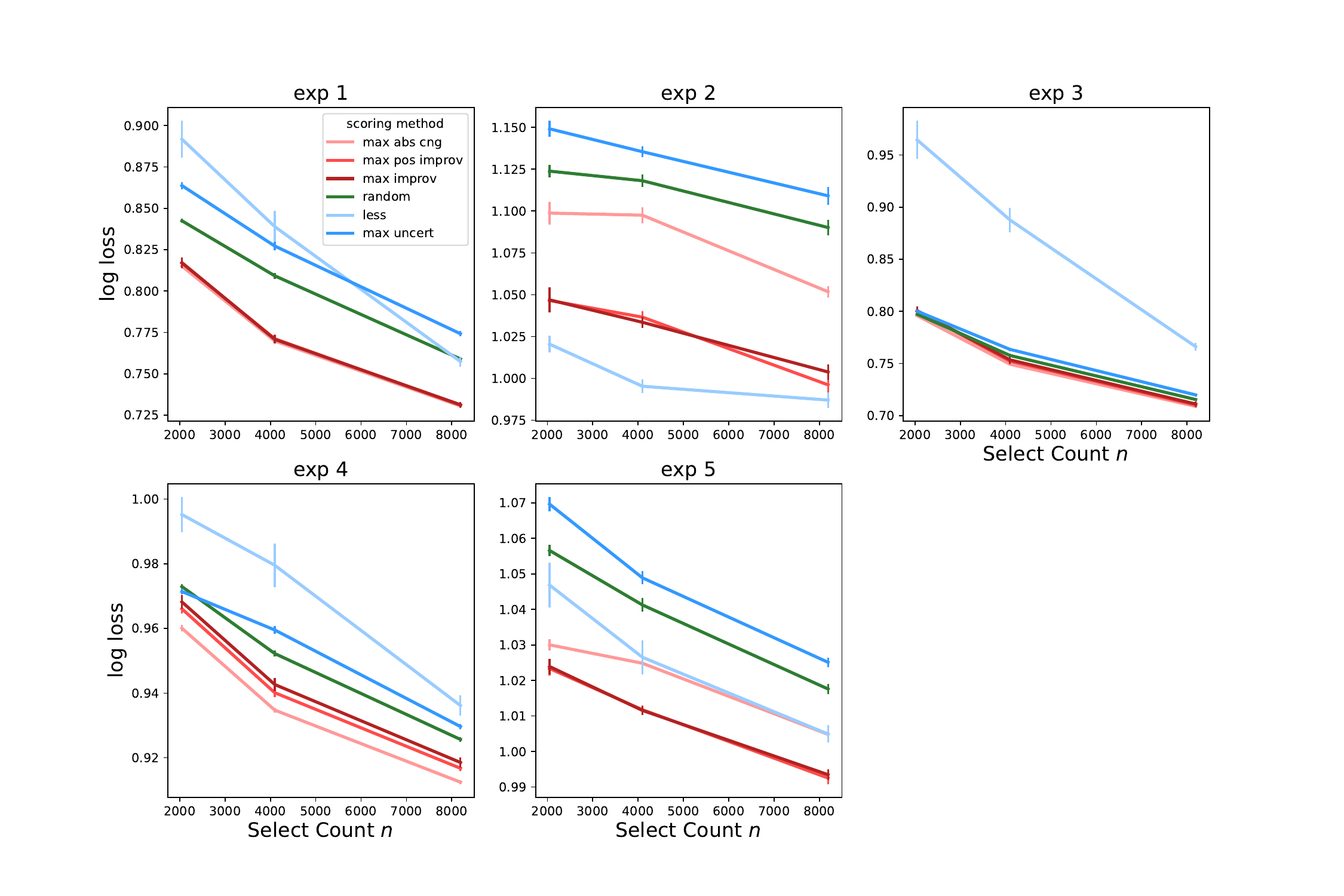}
\end{center}
    \caption{Expanded version of Figure~\ref{fig:IT_evol} including the Experiment~3 plot.}\label{fig:IT_evol_app}
\end{figure}

In the main paper, we compared our scoring methods for the \textsc{Score+Random} strategy. Due to space constraints, Figure~\ref{fig:IT_evol} omitted results for Experiment~3. In Figure~\ref{fig:IT_evol_app}, we provide an expanded version that includes results for Experiment~3 as well.

\subsection{Expanded Results for Named Entity Recognition}

In Figure~\ref{fig:NER_barplot} of the main paper, we reported results for \textsc{Score+Random} using a fixed selected sample size of \( n = 8 \times 1024 \), across all experiment configurations in Table~\ref{table:NER_exp}.

\begin{figure}[!htb]
\begin{center}
\includegraphics[trim=50 20 0 20,width=1.1\textwidth]
{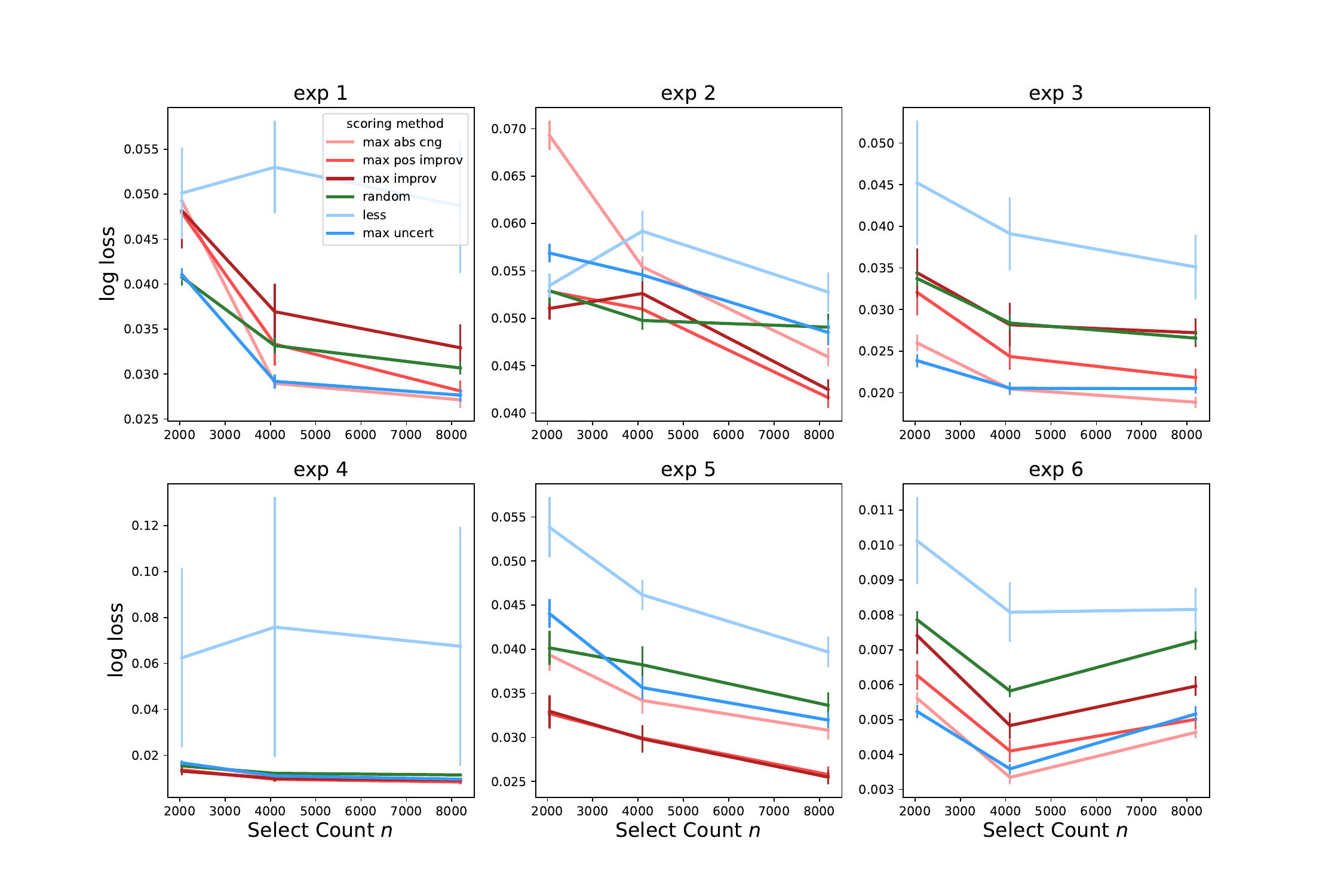}
\end{center}
    \caption{
    Test log-loss vs. number of selected samples $n$ for NER.
Lines show mean log-loss over 10 runs; error bars are $\pm$1 standard error.
Results use Method~A with the \textsc{Score+Random} strategy.
}\label{fig:NER_evol_app}
\end{figure}

In Figure~\ref{fig:NER_evol_app}, we show how the test log-loss varies with the selected sample size \( n \) for different scoring methods under the \textsc{Score+Random} strategy, and how these compare to random selection.

\section{Comparison of \textsc{Score+Random} and \textsc{Score-Only} selection}
In this section we examine how the performance of our strategies changes when all training examples are selected from the top-scoring set (\textsc{Score-Only}) instead of selecting only half of them from the top and the other half at random (\textsc{Score+Random}).

\begin{figure}[!htb]
\begin{center}
\includegraphics[width=\textwidth]{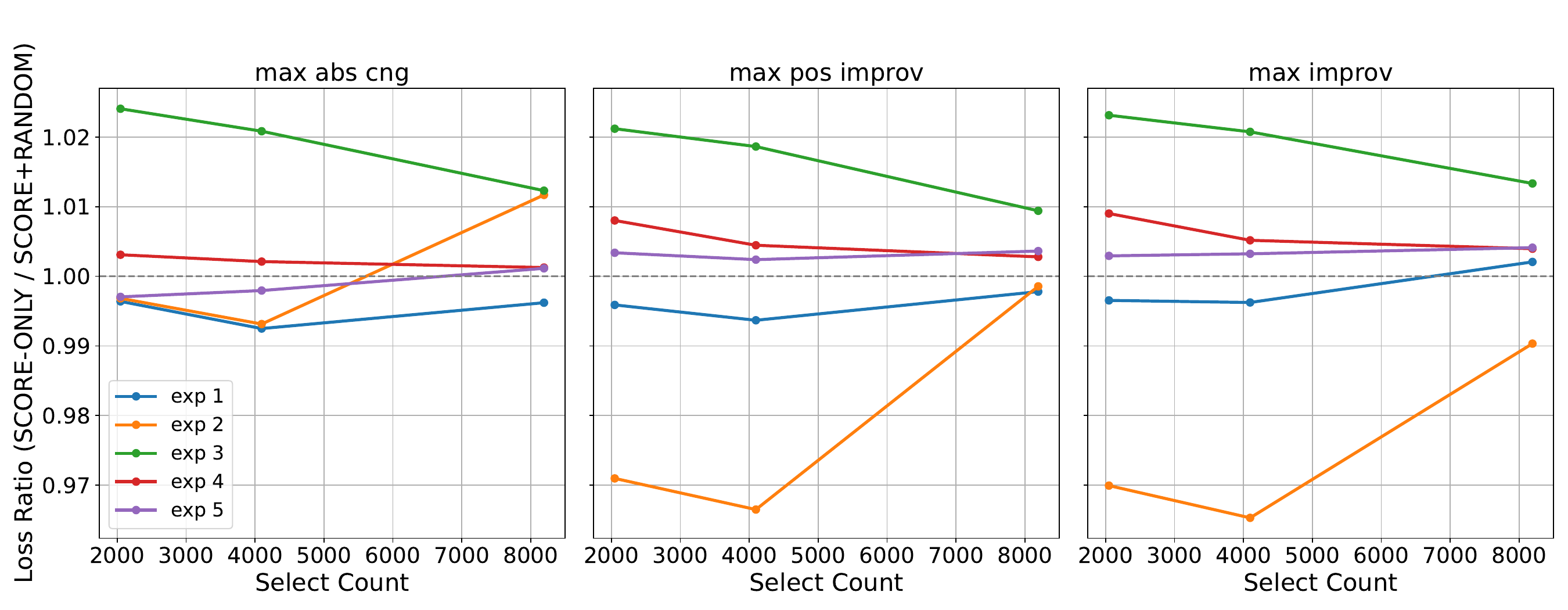}
\end{center}
    \caption{Ratio of log loss for \textsc{Score-Only} versus \textsc{Score+Random} across our three scoring strategies and all instruction-tuning setups in Table~\ref{table:IT_exp}. Scores are computed using Algorithm~\ref{alg:tov}.}\label{fig:ITselcomp}
\end{figure}

\begin{figure}[h]
\begin{center}
\includegraphics[width=\textwidth]{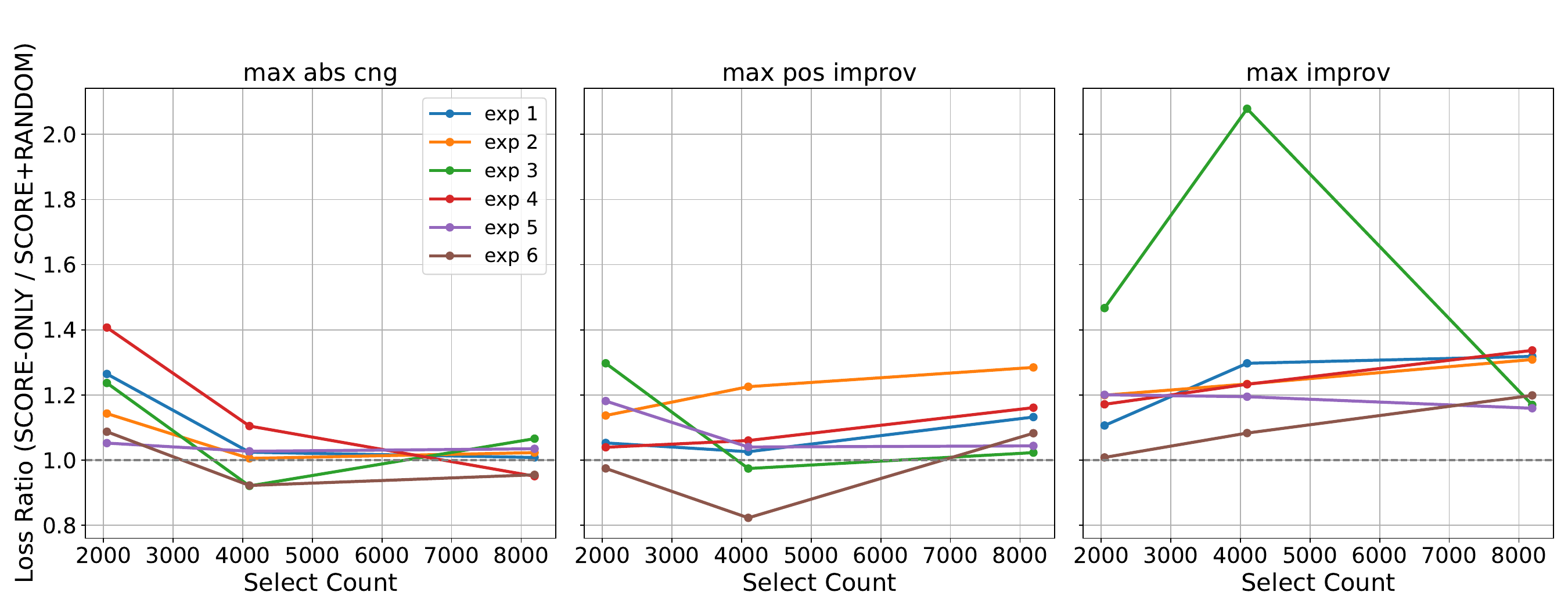}
\end{center}
  \caption{Ratio of log-loss for \textsc{Score-Only} versus \textsc{Score+Random} across our three scoring strategies and all NER setups in Table~\ref{table:NER_exp}.
Scores are computed using Algorithm~\ref{alg:tov}.}\label{fig:NERselcomp}
\end{figure}

Recall that our scores approximate how much benefit each example provides when added to a randomly chosen pool of training data.
A higher score therefore indicates an example expected to be more helpful in that setting.
\textsc{Score+Random} selects half of the final training set from the highest-scoring examples and fills the rest with random examples, whereas \textsc{Score-Only} takes only the top-scoring examples.
This design creates a trade-off:
\begin{itemize}
    \item Pure exploitation: Selecting only top-scoring examples can maximize immediate gain because every chosen example has a high estimated contribution.
    \item 
Score validity and diversity: The scores are defined relative to adding examples to a random pool.
If we select only top examples, the resulting set may differ substantially from the random reference, making the scores a less accurate guide.
Randomly adding half the examples keeps the final set closer to the conditions under which the scores were computed and also protects against loss of diversity.
\end{itemize}
Which effect dominates varies by task.

Figures~\ref{fig:ITselcomp} and~\ref{fig:NERselcomp} show the ratio of log-loss for the two selection strategies in instruction tuning and NER respectively.
In each figure the three subplots correspond to our three scoring strategies; different lines indicate the various experimental setups. Algorithm~\ref{alg:tov} is used to obtain the scores. 

For instruction tuning, \textsc{Score+Random} performs better in three setups (3, 4, 5), while \textsc{Score-Only} is better in the remaining two (1, 2) across most selection sizes and scoring methods.
For NER, \textsc{Score+Random} tends to outperform \textsc{Score-Only} more often, particularly for the Max-Improvement scores.

\section{Method B vs Method A}
\label{sec:BvsA}
All previous plots used Method~A (Algorithm~\ref{alg:tov}) for scoring.
Here we compare the performance of the two scoring methods: Method~A (Algorithm~\ref{alg:tov}) and Method~B (Algorithm~\ref{alg:tovB})—across our experiments, using the \textsc{Score+Random} selection strategy for both.

\begin{figure}[!htb]
\begin{center}
\includegraphics[width=\textwidth]{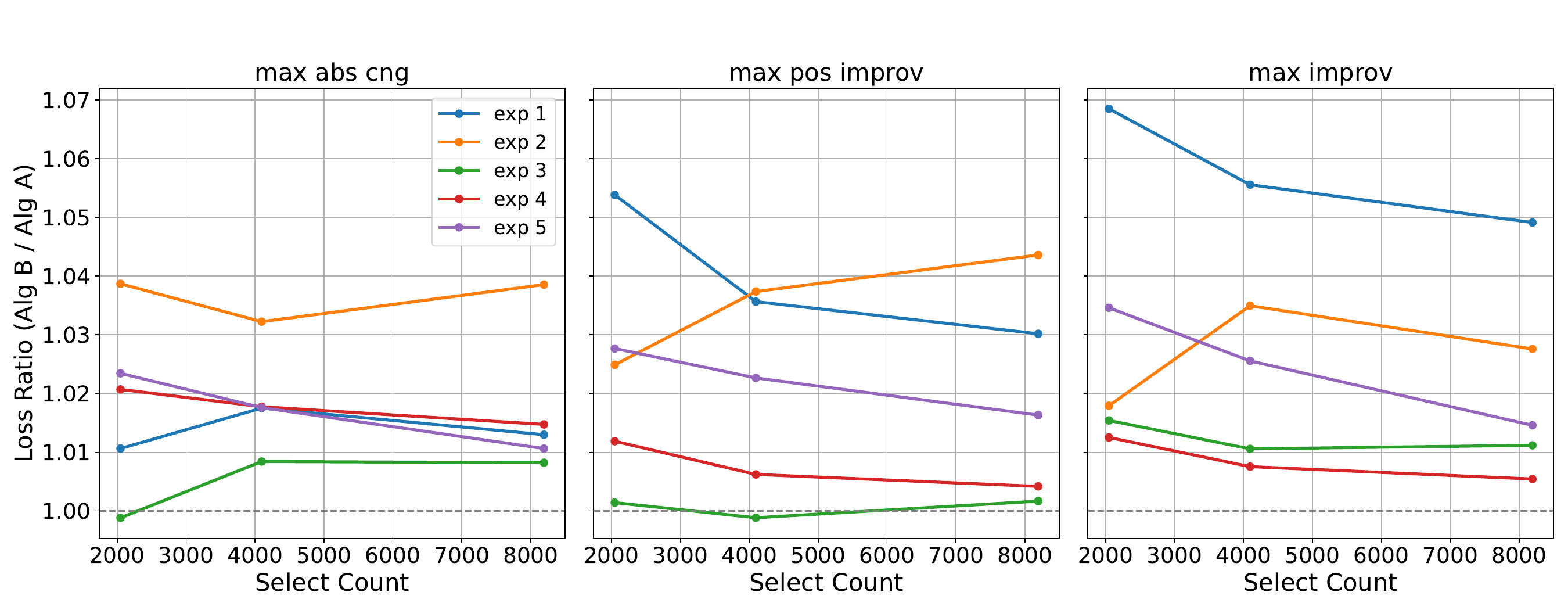}
\end{center}
    \caption{
    Ratio of test log-loss using Method~B (Algorithm~\ref{alg:tovB}) to Method~A (Algorithm~\ref{alg:tov}) for instruction tuning.
Results use the \textsc{Score+Random} selection strategy.
Each subplot corresponds to one scoring strategy; lines denote different experimental setups in Table~\ref{table:IT_exp}.
     }\label{fig:ITmet}
\end{figure}

\begin{figure}[!htb]
\begin{center}
\includegraphics[width=\textwidth]{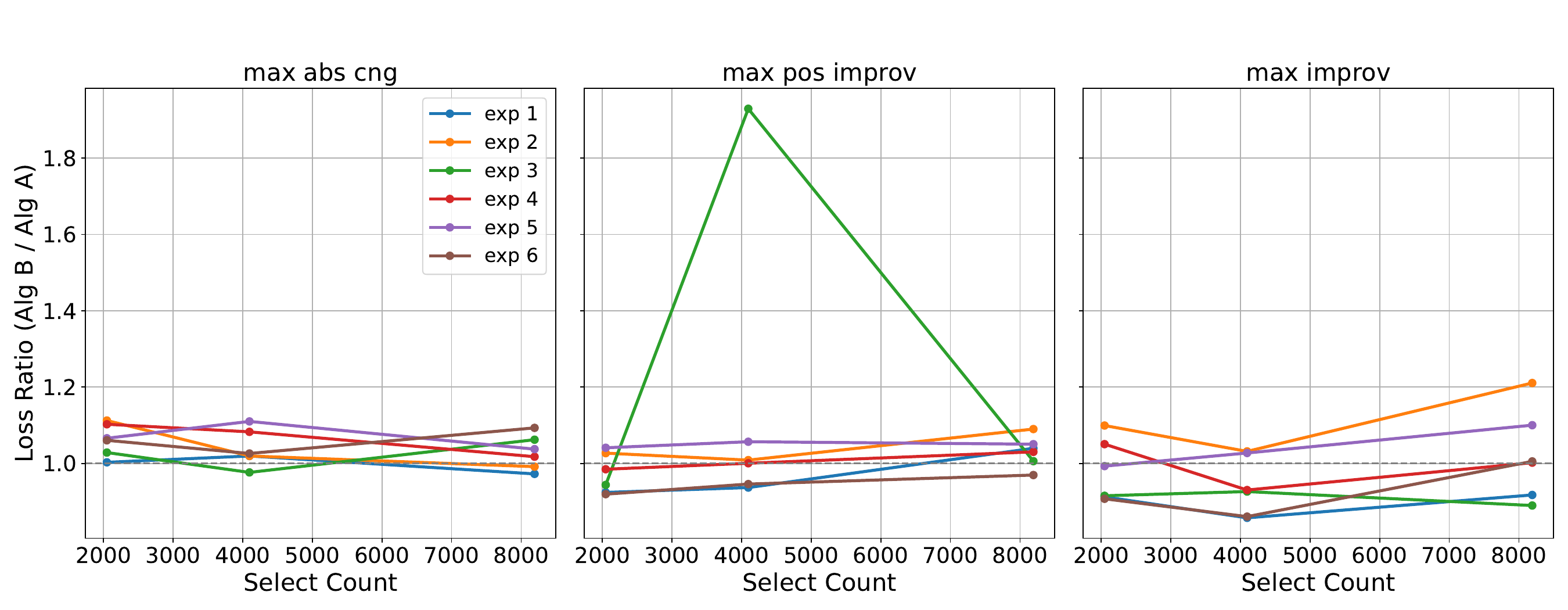}
\end{center}
    \caption{
    Ratio of test log-loss using Method~B (Algorithm~\ref{alg:tovB}) to Method~A (Algorithm~\ref{alg:tov}) for NER.
Results use the \textsc{Score+Random} selection strategy.
Each subplot corresponds to one scoring strategy; lines denote different experimental setups in Table~\ref{table:NER_exp}.
     }\label{fig:NERmet}
\end{figure}

Figures~\ref{fig:ITmet} and \ref{fig:NERmet} show the ratio of test log-loss obtained with Method~B relative to Method~A for instruction tuning and NER, respectively.
Each figure contains three subplots corresponding to our three scoring strategies, and different lines represent the various experimental setups.

The results indicate that for instruction tuning, Method~A is most often superior, while for NER there is no consistent winner.
A possible explanation is that Method~B uses two distinct training trajectories.
Our analysis assumes that the resulting models differ only slightly, but in practice, the two training trajectories can diverge substantially.
This effect is likely to be stronger with large and highly overparameterized models such as Meta-Llama-3-8B, which we used for instruction tuning, 
We expect the larger distance between the two models to result in
less accurate score estimation in Method~B, as compared to Method~A.

\section{Logistic Regression Experiments}\label{app:logreg}

In these experiments, we synthetically generated the training pool, validation set, and test set. We begin by defining a parametric family of distributions used to construct the data.

For a given \( p > 0 \) and parameter vector \( \theta \in \mathbb{R}^p \), we define a distribution \( \mathcal{P}_\theta \) over pairs \( (x, y) \), where \( x \in \mathbb{R}^p \) and \( y \in \{0,1\} \). The features are sampled as \( x \sim \mathcal{N}(0, I) \), and the label \( y \) is drawn according to a logistic model:
\[
\Pr(y = 1 \mid x) = \frac{1}{1 + \exp(-x \cdot \theta)}, \quad \Pr(y = 0 \mid x) = 1 - \Pr(y = 1 \mid x).
\]

We randomly sample a unit vector \( \theta^* \) from the unit sphere to serve as the target direction. A second unit vector \( \theta' \) is then drawn such that it lies at an angle \( \gamma \) from \( \theta^* \). In our experiments, we set \( p = 10 \) and \( \gamma = \pi/2 \).

The training pool consists of \( N = 128 \times 1024 \) samples, drawn independently from the mixture distribution:
\[
\mathcal{D}_{\text{train}} = \frac{1}{2} \mathcal{P}_{\theta^*} + \frac{1}{2} \mathcal{P}_{\theta'}.
\]

The validation and test sets contain \( m_{\sval} = 1024 \) and \( m_{\stst} = 10{,}000 \) samples respectively, both drawn i.i.d. from the target distribution \( \mathcal{P}_{\theta^*} \).

\begin{figure}[H]
\begin{center}
    \includegraphics[width=\textwidth]{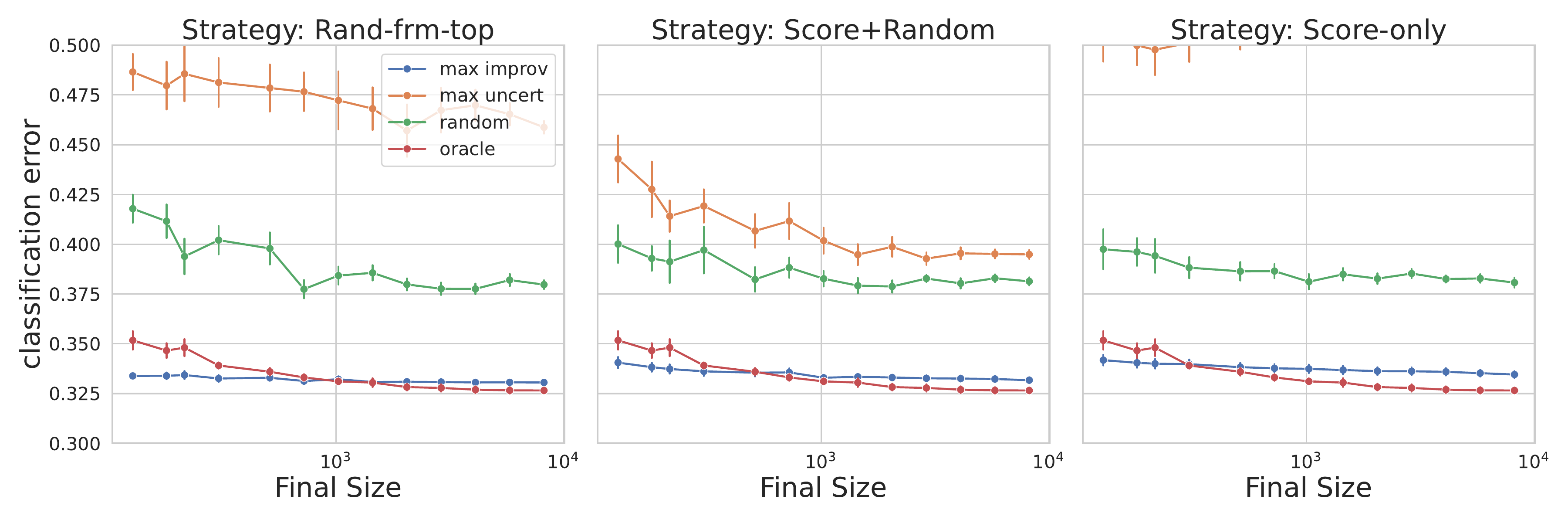}
\end{center}
    \caption{Data selection experiments with Method B for logistic regression on synthetic data in $d=10$ dimensions (4 epochs of training).
    Each color corresponds to a distinct method to score data in the training pool, and 
    each frame to a distinct method to use the score to form the selected set.
    Each symbol corresponds to the average of 10 experiments.}\label{fig:log_reg_exp}
\end{figure}

For scoring, we used Method B as described in Algorithm~\ref{alg:tovB}; similar experiments with Method~A produced comparable results, so we report only Method~B here. 
Algorithm \ref{alg:tovB} does not specify the method to select data on the basis of scores. 
In Figure \ref{fig:log_reg_exp} we compare {\sc score-only} and {\sc score$+$random} (already introduced above)
with a third one {\sc random-from-top} that selects at random from the top $50\%$ subset of data with highest scores. 

The \textsc{Random-from-Top} method is included only for these synthetic logistic-regression experiments, for theoretical interest as by construction, the training pool contains half of its examples from the target distribution.

The base set used for initial training contains \( |U| = 4 \times 1024 \) examples.

For both the scoring model and the final model training, we used 4 epochs of batch gradient descent with a linear decay learning rate scheduler. The initial learning rate was set to 0.5. We used \( \epsilon = \frac{1}{10} \) for adjusting the learning rate on validation examples.

The selected subset size \( n \) was varied from 128 to 8192 in multiplicative steps of \( \sqrt{2} \). All results are averaged over 10 independent runs. The final performance curves are presented in Figure~\ref{fig:log_reg_exp}.

\section{Proof of Proposition \ref{propo:InfluenceFunctions}}

Throughout this proof, we denote by $C$ a generic constant that can depend on 
$c_0, C_1, M, C_{\eta}$ and whose value is allowed to change from line to line.

Letting $\bDelta_{s}(i) =\hbtheta_{s}^{\ssur+i}-\hbtheta_{s}^{\ssur}$, Eq.~\eqref{eq:TwoGD}
yields
\begin{align}
\bDelta_{k+1}(i) & = \bDelta_{k}(i) -\eta m
\nabla^2 \hR_U(\hbtheta_{k}^{\ssur})  \bDelta_{k}(i) 
-\eta \nabla \ell(\hbtheta_{k}^{\ssur};\bx_i)+\err_k(i)\nonumber\\
 & = \bH_k\bDelta_{k}(i) 
-\eta \nabla \ell(\hbtheta_{k}^{\ssur};\bx_i)+\err_k(i)
\, ,\label{eq:TwoGD-Lin}
\end{align}
where $\bH_k$ is defined as in Eq.~\eqref{eq:MDef} and
\begin{align*}
\err_k(i) := -\eta\big[\nabla \ell(\hbtheta_{k}^{\ssur+i};\bx_i)-\nabla \ell(\hbtheta_{k}^{\ssur};\bx_i)\big] -\eta m
\int_0^1\big[\nabla^2 \hR_U(\obtheta_{k}(z))- \nabla^2 \hR_U(\hbts_{k})\big]\bDelta_{k}(i) \de z\, ,
\end{align*}
where $\obtheta_{k}(z)=(1-z)\hbts_{k}+z\hbtheta_{k}^{\ssur+i}$.
By assumption $\btheta\mapsto \nabla\ell(\btheta;\bx_i)$ and 
$\btheta\mapsto \nabla^2\hR_U(\btheta)$ are $M$-Lipschitz, whence
\begin{align}
\|\err_k(i)\|& \le \eta M \|\hbtheta_{k}^{\ssur+i}-\hbtheta_{k}^{\ssur}\| +\eta m M
 \|\hbtheta_{k}^{\ssur+i}-\hbtheta_{k}^{\ssur}\| \|\bDelta_{k}(i) \| \nonumber\\
 & = \eta M \|\bDelta_{k}(i)\| +\eta m M \|\bDelta_{k}(i) \|^2\, .\label{eq:ErrDelta}
\end{align}
Define $\bDelta^{\slin}_{k}(i)$ by letting $\bDelta^{\slin}_{k}(i)=\bzero$
and, for $k\ge 0$, 
\begin{align}
\bDelta^{\slin}_{k+1}(i) 
 & = \bH_k\bDelta^{\slin}_{k}(i) -\eta \nabla \ell(\hbtheta_{k}^{\ssur};\bx_i)\, .
 \label{eq:DeltaLinRecursion}
\end{align}
Comparing with Eq.~\eqref{eq:TwoGD-Lin}, we obtain
\begin{align}
\big(\bDelta_{k+1}(i)-\bDelta_{k+1}^{\slin}(i)\big) &= \bH_k \big(\bDelta_{k+1}(i)-\bDelta_{k+1}^{\slin}(i)\big)+\err_k(\eta,m)\, ,\nonumber\\
\Rightarrow\; \bDelta_{t}(i)-\bDelta_{t}^{\slin}(i) & =\sum_{s=0}^{t-1}\bM_{t,s+1}\err_s(\eta,m)\, .
\end{align}
Since $\nabla^2 \hR_U(\hbtheta_{k}^{\ssur})\succeq c_0\bI_d$,
we have $\|\bH_k\|_{\op}\le (1-c_0m\eta)$, and therefore
\begin{align}
 \|\bDelta_{t}(i)-\bDelta_{t}^{\slin}(i) \|& \le \sum_{s=0}^{t-1}\|\bM_{t,s+1}\|_{\op}
 \|\err_s(\eta,m)\|\nonumber\\
 & \le \sum_{s=0}^{t-1}\big(1-c_0m\eta\big)^{t-s-1} \|\err_s(\eta,m)\|\, .\label{eq:Delta_Delta}
 \end{align}
Further, from Eq.~\eqref{eq:DeltaLinRecursion}, 
and using $\|\nabla \ell(\hbtheta_{k}^{\ssur};\bx_i)\|\le C_1$, we get
\begin{align}
\bDelta_{t}^{\slin}(i) & =-\eta\sum_{s=0}^{t-1}\bM_{t,s+1}\nabla\ell(\hbtheta^{\ssur}_s;\bx_i)\, ,
\nonumber\\
 \Rightarrow\|\bDelta_{t}^{\slin}(i)\|&\le C_1\eta \sum_{s=0}^{t-1}\big(1-c_0m\eta\big)^{t-s-1} \le 
 \frac{C}{m}\,.\label{eq:Delta_lin}
\end{align}

Let $D_t(i) :=\max_{s\le t}\|\bDelta_s(i)\|$, $E_t(i):= \max_{s\le t}\|\err_s(i)\|$.
Using Eqs.~\eqref{eq:ErrDelta}, \eqref{eq:Delta_Delta} and \eqref{eq:Delta_lin}, we get
\begin{align*}
    D_t(i) &\le \frac{C}{m}+\frac{1}{c_0 m \eta}E_{t-1}(i)\, ,\\
    E_t(i) & \le \eta M D_t(i) +\eta m M D_t(i)^2\, .
\end{align*}
Using these inequalities together, we obtain, for all $m\ge m_0$
(and eventually adjusting the constant $C$)
\begin{align}
    D_t(i) &\le \frac{C}{m}\, ,\;\;\; E_t(i) \le \frac{C\eta}{m}\, ,
\end{align}
whence, using again Eq.~\eqref{eq:Delta_Delta}, we get
\begin{align}
 \|\bDelta_{t}(i)-\bDelta_{t}^{\slin}(i) \|& \le \frac{C}{m^2}\, .
 \end{align}

Notice that we can rewrite
\begin{align}
\score^{\slin}_i = -\frac{1}{L}\sum_{s=1}^L
\<\nabla \hR_{\sval}(\hbtheta_{s}^{\ssur}),\bDelta_s^{\slin}(i)\>\, ,
\end{align}
whence, using the fact that $\|\nabla^2 \hR_{\sval}(\obtheta)\|_{\op}\le C$
for all $\obtheta \in [\hbts_{k},\hbtheta_{k}^{\ssur+i}]$
(this follows from the assumed bound $\|\nabla^2 \hR_{\sval}(\hbtheta_{k}^{\ssur})\|_{\op}\le C_1$
and the Lipschitz property of $\btheta\mapsto \nabla^2 \hR_{\sval}(\hbtheta)$), we get
\begin{align*}
\big|\score_i-\score^{\slin}_i\big| &\le C\max_{s\le L}\|\bDelta_s(i)\|^2
+\frac{1}{L}\sum_{s=1}^L
\big|\<\nabla \hR_{\sval}(\hbtheta_{s}^{\ssur}),\bDelta_s(i)-\bDelta^{\slin}_s(i)\>|\\
& \le C \, \max_{s\le L}\|\bDelta_s(i)\|^2 + C \max_{s\le L}\|\bDelta_s(i)-\bDelta^{\slin}_s(i)\|\\
&\le \frac{C}{m^2}\, ,
\end{align*}
and this completes the proof. 
%
%
\section{Proof of Theorem \ref{propo:Score}}

Throughout this proof, we denote by $C$ a generic constant that can depend on 
$c_0, C_1, M, C_{\eta}$ and whose value is allowed to change from line to line.

\subsection{Bound on $\Upsilon_i$}

The iteration for $\hbts_k$ and $\hbtss_k$, as specified by Algorithm \ref{alg:tovB},
reads
\begin{align}
\hbts_{k+1} &= \hbts_{k} -\eta m\nabla \hR_U(\hbts_{k})\, ,\\
\hbtss_{k+1} &= \hbtss_{0,k+1} -\eps\eta \mval\nabla \hR_{\sval}(\hbtss_{0,k+1})\, ,\;\;\;\;\;\;
\hbtss_{0,k+1} = \hbtss_k - \eta m \nabla\hR_U(\hbtss_{k})\, .
\end{align}
Letting $\bDelta_k:= \hbtss_{k} -\hbts_{k}$, and $\bDelta_{0,k}:= \hbtss_{0,k} -\hbts_{k}$,
we obtain
\begin{align}
\bDelta_{k+1} &=  \bDelta_{0,k+1} - \eps\eta \mval\nabla\hR_{\sval}(\hbts_{k+1})  +\errf_{k+1}\, ,\label{eq:Reck}\\
\bDelta_{0,k+1} & = \bH_k\bDelta_k  +\errs_k\, .\label{eq:Rec0k}
\end{align}
where, letting $\obtheta_{0,k+1}(z) = (1-z) \hbts_{k+1}+z \hbtss_{0,k+1}$
and $\obtheta_{k}(z) = (1-z) \hbts_{k}+z \hbtss_{k}$, we have
\begin{align*}
\errf_{k+1} & := -\eta \eps \mval\int_0^1\nabla^2 \hR_{\sval}(\obtheta_{0,k+1}(z)) 
\bDelta_{0,k+1} \, \de z \, ,\\
\errs_k &:= - \eta m\int_0^1\big[\nabla^2\hR_{U}(\obtheta_{k}(z))-
\nabla^2\hR_{U}(\hbts_{k}) \big]\bDelta_k\, \de z\, .
\end{align*}
Using the assumption that $\|\nabla^2 \hR_{\sval}(\hbts_{k+1}(z))\|_{\op}\le C$ and
$\btheta\mapsto \nabla^2 \hR_{\sval}(\btheta)$ is $M$-Lipschitz, we get:
\begin{align}
\|\errf_{k+1}\|&\le C\eps\eta \mval  \big\{\|\bDelta_{0,k+1}\|+  \|\bDelta_{0,k+1}\|^2\big\}\, .
\end{align}
On the other hand,  since $\btheta\mapsto \nabla^2 \hR_{U}(\btheta)$ is also $M$-Lipschitz, we have 
\begin{align}
  \|\errs_k\|&\le \eta m M\|\bDelta_k\|^2\, ,\label{eq:ErrsBound}
\end{align}
whence, using Eq.~\eqref{eq:Rec0k} and $\|\bH_k\|_{\op}\le 1$
\begin{align}
    &\|\bDelta_{0,k+1}\| \le \|\bDelta_k\|  + \eta m M\|\bDelta_k\|^2\nonumber\\
   & \Rightarrow \|\errf_{k+1}\|\le C\eps\eta \mval
   \big\{ \|\bDelta_k\|  + \|\bDelta_k\|^2 + \eta^2 m^2 \|\bDelta_k\|^4\big\}\, , 
   \label{eq:ErrfBound}
\end{align}
where in the last line we used the assumption that $\eta m\le C_{\eta}$.

Substituting Eqs.~\eqref{eq:ErrsBound} and \eqref{eq:ErrfBound} in Eq.~\eqref{eq:Reck}, \eqref{eq:Rec0k}, 
we obtain (using again $\eta m\le C_{\eta})$
\begin{align}
\bDelta_{k+1} &= \bH_k \bDelta_k - \eps\eta \mval \nabla\hR_{\sval}(\hbts_{k+1}) +\err_k\, ,\\
\|\err_k\| & \le  C\eta \eps \mval \big(\|\bDelta_k\|  + \|\bDelta_k\|^4\big)
+ C\eta m \|\bDelta_k\|^2\, .\label{eq:Err-Delta}
\end{align}

We define  $\bDelta^{\slin}_k=\bzero$ and, for $k\ge 0$,
\begin{align}
\bDelta^{\slin}_{k+1} &= \bH_k \bDelta^{\slin}_k - \eps\eta\mval \nabla\hR_{\sval}(\hbts_{k+1})\, ,
\end{align}
whence
\begin{align}
\bDelta^{\slin}_{t} &= -\eps \eta \mval \sum_{s=0}^{t-1}\bM_{t,s+1}  \nabla\hR_{\sval}(\hbts_{s+1}) \, ,\;\;\;\;\;
\bDelta_{t}-\bDelta^{\slin}_{t} =\sum_{s=0}^{t-1}\bM_{t,s+1}  \err_s\, .\label{eq:Err-Delta-2}
\end{align}

Define $D_t:= \max_{s\le t}\|\bDelta_s\|$, $E_t:=\max_{s\le t}\|\err_s\|$. 
Using the fact that $\|\bM_{t,s+1}\|_{\op}\le (1-c_0m\eta)^{t-s-1}$ 
and the assumption $\|\nabla\hR_{\sval}(\hbts_k)\|\le C_1$, we get,
from Eqs.~\eqref{eq:Err-Delta}, \eqref{eq:Err-Delta-2},
\begin{align}
D_{t+1} &\le \frac{C\eps\mval}{m} +\frac{C}{m\eta} E_t\, ,\\
E_t & \le C\eta\eps \mval (D_t+D_t^4)  +C\eta m D_t^2\, ,
\end{align}
Using the assumption $\eps m_{\sval}/m\le c_*$, this is easily seen to imply
\begin{align}
D_{t} &\le C\frac{\eps\mval}{m}\, ,\;\;\;\;\;\;\;\;\; E_t  \le C
\frac{(\eps\mval)^2}{m}\eta\, .
\end{align}
Substituting in Eq.~\eqref{eq:Err-Delta-2}, we get
\begin{align}
\|\bDelta_{t}-\bDelta^{\slin}_{t}\|\le \sum_{s=0}^{t-1}(1-c_0m\eta)^{t-s-1} \| \err_s\|\le
C\Big(\frac{\eps m_{\sval}}{m}\Big)^2\, .
\end{align}

The linearized score of Eq.~\eqref{eq:PhiLin} can be rewritten as
\begin{align}
\Upsilon_i^{\slin}\eps = 
-\frac{1}{L}\sum_{s=1}^{L}\<\nabla\ell(\hbts_s;\bx_i),\bDelta_s^{\slin}\>\, .
\end{align}
Using the fact that $\|\nabla\ell(\hbts_k;\bx_i)\|, \|\nabla\ell(\hbts_k;\bx_i)\|_{\op}\le C_1$,
we get
\begin{align}
\big|\Upsilon_i(\eps)- \Upsilon_i^{\slin}\eps \big| &\le \frac{1}{L}\sum_{s=1}^{L}
\Big|\ell(\hbts_k; \bx_i)-\ell(\hbtss_k; \bx_i)+
\<\nabla\ell(\hbts_s;\bx_i),\bDelta_s^{\slin}\>\Big|\\
&\le \frac{C}{L}\sum_{s=1}^{L}\|\bDelta_s\|^2+ \frac{C}{L}\sum_{s=1}^{L}\|\bDelta_s-\bDelta_s^{\slin}\|\\
&\le C \Big(\frac{\eps m_{\sval}}{m}\Big)^2\, ,
& 
\end{align}

%
%

\subsection{Bound on  $\phi_i$}

The iteration for $\hbts_k$ and $\hbtss_k$, as specified by Algorithm \ref{alg:tovB},
reads
\begin{align}
\hbts_{k+1} &= \hbts_{k} -\eta m\nabla \hR_U(\hbts_{k})\, ,\\
\hbtv_{k+1} &= \hbts_{k+1} -\eps\eta \mval\nabla \hR_{\sval}(\hbts_{k+1})\,  .
\end{align}
Hence, we can rewrite 
\begin{align*}
\phi^{\slin}_i\eps = -\frac{1}{L}\sum_{s=1}^{L}\<\nabla\ell(\hbts_s;\bx_i),\hbtv_{s} - \hbts_{s} 
\>\, .
\end{align*}
Using the assumptions $\|\nabla^2\ell(\hbts_s;\bx_i)\|_{\op}\le C_1$,
$\|\hR_{\sval}(\hbts_{k})\|\le C_1$, we obtain
\begin{align}
    \big|\phi_i(\eps)-\phi^{\slin}_i\eps\big|&\le 
    \frac{1}{L}\sum_{s=1}^{L}\Big|\ell(\hbtv_s;\bx_i)-\ell(\hbts_s;\bx_i)-
    \<\nabla\ell(\hbts_s;\bx_i),\hbtv_{s} - \hbts_{s} \>
    \Big|\\
    &\le 
    \frac{C}{L}\sum_{s=1}^{L}\|\hbtv_{s} - \hbts_{s} \|^2 \le C(\eps\eta\mval)^2\, .
\end{align}
The claim thus follows by recalling that $\eta\le C_{\eta}/m$.
%
%
\section{Proof of Theorem \ref{thm:LargeL_cvx}}

To lighten notation, we define $\br_k := \nabla \hR_{\sval}(\hbts_k)$
and $\bv_k(i):=\nabla \ell(\hbts_k;\bx_i)$.

For any $L_,L_1\in\integers$, we have 
\begin{align*}
\Upsilon^{\slin}_i(L) &=\Upsilon^{\slin,0}_i(L)+\Upsilon^{\slin,1}_i(L)+\Upsilon^{\slin,2}_i(L)
+\Upsilon^{\slin,3}_i(L)\, ,\\
\Upsilon^{\slin,0}_i(L) &:=
\frac{\eta\mval}{L}\sum_{0\le t<s\le L_0}
\<\br_{t+1},\bM_{s,t+1}^{\sT}\bv_s(i)\>\, ,\\
\Upsilon^{\slin,1}_i(L) &:=
\frac{\eta\mval}{L}\sum_{0\le t\le L_0, L_0<s\le L}
\<\br_{t+1},\bM_{s,t+1}^{\sT}\bv_s(i)\>\, ,\\
\Upsilon^{\slin,2}_i(L) &:=
\frac{\eta\mval}{L}\sum_{L_0< t<s\le L_0: |s-t|\ge L_1}
\<\br_{t+1},\bM_{s,t+1}^{\sT}\bv_s(i)\>\, ,\\
\Upsilon^{\slin,2}_i(L) &:=
\frac{\eta\mval}{L}\sum_{L_0< t<s\le L_0: |s-t|\ge L_1}
\<\br_{t+1},\bM_{s,t+1}^{\sT}\bv_s(i)\>\, ,\\
\Upsilon^{\slin,3}_i(L) &:=
\frac{\eta\mval}{L}\sum_{L_0< t<s\le L_0: |s-t|<L_1}
\<\br_{t+1},\bM_{s,t+1}^{\sT}\bv_s(i)\>\, .
\end{align*}
Since by continuity we have $\lim_{k\to\infty}\nabla^2 \hR_U(\hbts_{k})= \bQ_{\infty}$,
for any $\delta\in (0,1/2)$, we can choose $L_0$ large enough so that $(1-\delta) \bQ_{\infty}  \preceq\nabla^2 \hR_U(\hbts_{k})\preceq(1+\delta) \bQ_{\infty}$ for all $k>L_0$.
In particular there exists $c_0>0$ (independent of $\eps$) 
such that $\|\bH_k\|_{op}\le (1-c_0m\eta)$ for 
all $k>L_0$.

Clearly $|\Upsilon^{\slin,0}_i(L)|\le C(L_0)/L\to 0$ as $L\to\infty$.
Further 
\begin{align*}
\big|\Upsilon^{\slin,1}_i(L)\big| &\le \frac{C\eta\mval}{L}\sum_{0\le t\le L_0, L_0<s\le L}
\|\bM_{s,t+1}\|_{\op} \\
&\le \frac{C\eta\mval}{L}\sum_{0\le t\le L_0, L_0<s\le L}
(1-c_0m\eta)^{s-t-1}\\
&\le \frac{C\eta\mval}{L}\frac{L_0}{c_0m\eta} \to 0\, .
\end{align*}
%
%
%
%

Finally, by increasing  $L_0$, we can ensure that, for $k>L_0$,
$\|\bH_k-\bH_{\infty}\|_{\op}\le \delta$, $\|\br_k-\br_{\infty}\|\le \delta$,
$\|\bv_k(i)-\bv_{\infty}(i)\|\le \delta$ (where $\bH_{\infty}=\bI-\eta m\bQ_{\infty}$
and $\br_{\infty}$, $\bv_{\infty}(i)$). Hence
\begin{align*}
\big|\<\br_{t+1},\bM_{s,t+1}^{\sT}\bv_s(i)\>- \<\br_{\infty},\bH_{\infty}^{s-t-1}\bv_{\infty}(i)\>\big|
\le C|t-s+1|(1-c_0m\eta)^{s-t-1}\delta \, .
\end{align*}
Therefore, letting
\begin{align}
\tilde \Upsilon^{\slin,2}_i(L) &:=
\frac{\eta\mval}{L}\sum_{L_0< t<s\le L}
 \<\br_{\infty},\bH_{\infty}^{s-t-1}\bv_{\infty}(i)\>\, ,
\end{align}
we have
\begin{align*}
\big|\Upsilon^{\slin,2}_i(L)-\tilde \Upsilon^{\slin,2}_i(L) \big|&\le 
\frac{\eta\mval}{L}\sum_{L_0< t<s\le L}
\big|\<\br_{t+1},\bM_{s,t+1}^{\sT}\bv_s(i)\>- \<\br_{\infty},\bH_{\infty}^{s-t-1}\bv_{\infty}(i)\>\big|\\
&\le 
\frac{\eta\mval}{L}\sum_{L_0< t<s\le L}  C|t-s+1|(1-c_0m\eta)^{s-t-1}\delta\\
&\le \eta\mval\cdot \frac{1}{(c_0m\eta)^2}\delta\, .
\end{align*}

Finally, using again  $|\<\br_{\infty},\bH_{\infty}^{s-t-1}\bv_{\infty}(i)\>|\le (1-c_0m\eta)^{s-t-1}$, we have
\begin{align*}
\lim_{L\to\infty}\tilde \Upsilon^{\slin,2}_i(L) &=\lim_{L\to\infty}\frac{\eta\mval}{L}\sum_{L_0< t\le L}\sum_{s=t+1}^{\infty}
 \<\br_{\infty},\bH_{\infty}^{s-t-1}\bv_{\infty}(i)\>\\
 &=\eta\mval\sum_{k=0}^{\infty}
 \<\br_{\infty},\bH_{\infty}^{k}\bv_{\infty}(i)\>\\
 &=\eta\mval
 \<\br_{\infty},(\bI-\bH_{\infty})^{-1}\bv_{\infty}(i)\>\\
 & = \frac{\mval}{m}
 \<\br_{\infty},\bQ_{\infty}^{-1}\bv_{\infty}(i)\>\, .
 \end{align*}

 This finishes the proof of the part of Eq.~\eqref{eq:InfluenceLocConvex}
 which concerns the limit of $\Upsilon_i^{\slin}$. 
 The calculation of $\lim_{L\to\infty} \score^{\slin}_i(L)$ is completely analogous and 
 we omit it.
%
%
\section{Proof of Theorem \ref{thm:LargeL-Linear}}

To simplify notations, we write $y_j=y(\bx_j)$ for the response variables and 
$\bpsi_j = \bpsi(\bx_j)$ for the feature vectors. Similarly, for the 
$y_j^{\sval}=y(\bz_j^{\sval})$, $\bpsi(\bz_j^{\sval}) =\bpsi_j^{\sval}$.

With these notations, we have $\bH_k=\bH$ independent of $k$ and
\begin{align}
\nabla\ell(\hbtheta;\bx_i) &= -(y_i-\<\bpsi_i,\btheta\>)\bpsi_i\, ,\\
\nabla\hR_{\sval}(\btheta) &=-\frac{1}{m} \bPsi^{\sT} \big(\by-\bPsi\btheta\big)\, ,\\
\bH & = \bI -\eta \bPsi^{\sT}\bPsi\, .
\end{align}
Hence, 
\begin{align}
\Upsilon^{\slin}_i &= \frac{\eta\mval}{L}\sum_{0\le t<s\le L}
r_s(i)\<\bPsi\br^{\sval}_{t+1} ,
\bH^{s-t-1}\bpsi_i\> \, ,\\
\br^{\sval}_t & := \by^{\sval}-\bpsi^{\sval}\hbts_{t+1}\, ,\;\;\;\;\;\;\;
r_s(i):= y_i-\<\bpsi_i,\hbts_s\>
\end{align}

Since $\bP_{\bPsi}$ is the projector onto the null-space of $\bH$,
and by our choice of $\eta$,
we have $\bH = \bP_{\bPsi}+\bH_{\perp}$, where the row/column space of $\bH_{\perp}$
is orthogonal to the one of $\bP_{\bPsi}$ and $\|\bH_{\perp}\|_{\op}=(1-c_{\psi}\eta)\in [0,1)$.
As a consequence $\|\bH^{s-t-1}-\bP_{\bPsi}\|_{\op}\le (1-c_{\psi}\eta)^{s-t-1}$.

Define
\begin{align}
\tilde\Upsilon^{\slin}_i &:= \frac{\eta\mval}{L}\sum_{0\le t<s\le L}
r_s(i)\<\bPsi\br^{\sval}_{t+1} ,
\bP_{\bPsi}\bpsi_i\> \, .
\end{align}
Then we have 
\begin{align*}
\Big|\frac{1}{L}\Upsilon^{\slin}_i - \frac{1}{L}\tilde\Upsilon^{\slin}_i\Big| &\le 
\frac{\eta\mval}{L^2}\sum_{0\le t<s\le L}
\Big|r_s(i)\<\bPsi\br^{\sval}_{t+1} ,
\bP_{\bPsi}\bpsi_i\> \Big|\\
&\le 
\frac{\eta\mval}{L^2}\sum_{0\le t<s\le L}
|r_s(i)|\, \|\bPsi\br^{\sval}_{t+1}\|
\|\bH^{s-t-1}-\bP_{\bPsi}\|_{\op}\ \big|\bpsi_i\big|\\
&\stackrel{(a)}{\le} C\frac{\eta\mval}{L^2}\sum_{0\le t<s\le L}(1-c_{\psi}\eta)^{s-t-1}\\
& \le C\frac{\eta\mval}{L} \frac{1}{c_{\psi}\eta}\stackrel{L\to\infty}{\longrightarrow} 0\, .
\end{align*}
In $(a)$, we used the fact that $\lim_{t\to\infty}\hbts_t = \hbtheta$
\citep{bartlett2021deep},
and therefore $|r_s(i)|$, $\|\bPsi\br^{\sval}_{t+1}\|$ remain bounded as $s,t\to\infty$. 

In view of the above,  $\lim_{L\to\infty}\Upsilon^{\slin}_i/L = \lim_{L\to\infty}\tilde\Upsilon^{\slin}_i/L$. For the latter, we have
\begin{align*}
\lim_{L\to\infty}
\frac{1}{L}\tilde\Upsilon^{\slin}_i &= \lim_{L\to\infty}\frac{\eta\mval}{L^2}\sum_{0\le t<s\le L}
r_s(i)\<\bPsi\br^{\sval}_{t+1} ,
\bP_{\bPsi}\bpsi_i\> \\
&= \lim_{L\to\infty}\frac{\eta\mval}{L^2}\sum_{L_0\le t<s\le L}
r_s(i)\<\bPsi\br^{\sval}_{t+1} ,
\bP_{\bPsi}\bpsi_i\> \\
& = \lim_{L\to\infty}\frac{\eta\mval}{L^2}\sum_{L_0\le t<s\le L}
r(i)\<\bPsi\br^{\sval} ,
\bP_{\bPsi}\bpsi_i\> +\err(L_0,L)\, ,
\end{align*}
where
\begin{align}
    |\err(L_0,L)|\le C\sup_{s\ge L_0}\|r_s(i)-r(i)|+ C\sup_{t\ge L_0}\|\br^{\sval}_{t}-\br^{\sval}_{t+1}|\, .
\end{align}
Since  $\lim_{t\to\infty}\hbts_t = \hbtheta$, we have
$\lim_{L_0\to\infty}\lim\sup_{L\to\infty}\err(L_0,L) = 0$.
Therefore, 

\begin{align*}
\lim_{L\to\infty}
\frac{1}{L}\tilde\Upsilon^{\slin}_i  &=
 \lim_{L_0\to\infty}\lim_{L\to\infty}\frac{\eta\mval}{L^2}\sum_{L_0\le t<s\le L}
r(i)\<\bPsi\br^{\sval} ,\bP_{\bPsi}\bpsi_i\>\\
&= 
\frac{1}{2}\eta\mval r(i)\<\bPsi\br^{\sval} ,\bP_{\bPsi}\bpsi_i\> \, .
\end{align*}
This proves the limit for $\Upsilon^{\slin}_i(L)$ in Eq.~\eqref{eq:LimitOverparam}.

The limit of $S^{\slin}_i(L)$ is computed essentially by the same argument and we omit the derivation.

%
%

\section{Limitation}

The core idea of ``train-on-validation" impacting training examples is general, but the specific scoring function $F(.)$ and aggregation strategy might need adaptation for different problem settings.

The \textsc{Score+Random} selection strategy often outperformed \textsc{Score-Only} in our experiments, suggesting that diversity plays an important role beyond simply selecting the ``most affected" examples. While this is a practical improvement, it also indicates that our current scoring mechanism might not fully capture the optimal diversity or coverage needed for effective generalization. It will be interesting to explore more sophisticated diversity-aware scoring or selection mechanisms that explicitly balance our scoring methods with representation across the data space. 

Although we mitigated bias toward shorter examples through length-based binning, a more refined length-normalization or task-specific weighting might further enhance the selection process.
Furthermore, it will be interesting to see if the performance of our strategies further improves compared to random selection if the learning rate is also tuned for these strategies and not just for random selection.

Finally, our theoretical analysis relies on stylized settings that are plausible for simple models but may not hold in many large-scale applications.

\section{Models and Datasets information}
\label{sec:datasets}

\subsection{Dataset information}
\begin{itemize}
\item Slim Orca:
\begin{itemize}
 \item \href{https://huggingface.co/datasets/Open-Orca/SlimOrca}{Link}
 \item Citations-\cite{longpre2023flan, mukherjee2023orca, SlimOrca}
\item Licence:  mit
\end{itemize}
\item Alpaca GPT-4:
\begin{itemize}
 \item {Paper}:\cite{peng2023instruction}
 \item  \href{https://github.com/Instruction-Tuning-with-GPT-4/GPT-4-LLM}{Repository}
 \item \href{https://huggingface.co/datasets/vicgalle/alpaca-gpt4}{Link}
\item Licence:  cc-by-nc-4.0
\end{itemize}
\item Alpaca GPT-3.5:
\begin{itemize}
\item Paper: \cite{alpaca}
 \item \href{https://huggingface.co/datasets/tatsu-lab/alpaca/blob/main/README.md}{Link}
\item Licence: cc-by-nc-4.0
\end{itemize}

\item Multinerd:
\begin{itemize}
\item Paper: \cite{tedeschi-navigli-2022-multinerd} 
 \item \href{https://huggingface.co/datasets/Babelscape/multinerd/blob/main/README.md}{Link}
\item Licence: cc-by-nc-sa-4.0
\end{itemize}

\item Ai4p:
\begin{itemize}
 \item \href{https://huggingface.co/datasets/ai4privacy/pii-masking-300k/tree/main}{Link}
\item Licence:  \href{https://huggingface.co/datasets/ai4privacy/pii-masking-300k/blob/main/LICENSE.md}{link}
\end{itemize}

\item C4 dataset:
\begin{itemize}
 \item \href{https://github.com/allenai/allennlp/discussions/5056}{Link}
 \item Labeled for NER task using llms.
\item Licence: \href{https://commoncrawl.org/terms-of-use}{terms of use}
\end{itemize}

\item Syn-Big:
\begin{itemize}
\item Synthetically generated by us using llms.
\item Proprietary dataset
\end{itemize}
\end{itemize}
\subsection{Pretrained Model information}
\begin{itemize}
    \item Meta-Llama-3-8B~\cite{llama3modelcard}
    \begin{itemize}
        \item \href{https://github.com/meta-llama/llama3/blob/main/MODEL_CARD.md}{Link}
        \item License: llama3
    \end{itemize}
    \item  xlm-roberta-base~\cite{DBLP:journals/corr/abs-1911-02116}
    \begin{itemize}
    \item  \href{https://huggingface.co/FacebookAI/xlm-roberta-base/blob/main/README.md}{Link}
        \item License: mit
    \end{itemize}

\end{itemize}

\end{document}